\documentclass[twoside]{IEEEtran}
\usepackage[T1]{fontenc}
\usepackage[latin9]{inputenc}
\usepackage{color}
\usepackage{float}
\usepackage{amsmath}
\usepackage{amssymb}
\usepackage{graphicx}

\makeatletter

\floatstyle{ruled}
\newfloat{algorithm}{tbp}{loa}
\providecommand{\algorithmname}{Algorithm}
\floatname{algorithm}{\protect\algorithmname}

\usepackage{cite}

\@ifundefined{showcaptionsetup}{}{%
 \PassOptionsToPackage{caption=false}{subfig}}
\usepackage{subfig}
\makeatother

\begin{document}

\title{Joint Sensing Matrix and Sparsifying Dictionary Optimization for
Tensor Compressive Sensing}

\author{Xin Ding, \textit{Student Member, IEEE}, Wei Chen, \textit{Member,
IEEE}, and Ian J. Wassell \thanks{Xin Ding and Ian J. Wassell are with the Computer Lab, University
of Cambridge, UK (e-mail: xd225, ijw24@cam.ac.uk).}\thanks{Wei Chen is with the State Key Laboratory of Rail Traffic Control
and Safety, Beijing Jiaotong University, China, and also with the
Computer Lab, University of Cambridge, UK (e-mail: wc253@cam.ac.uk).}}
\maketitle
\begin{abstract}
Tensor Compressive Sensing (TCS) is a multidimensional framework of
Compressive Sensing (CS), and it is advantageous in terms of reducing
the amount of storage, easing hardware implementations and preserving
multidimensional structures of signals in comparison to a conventional
CS system. In a TCS system, instead of using a random sensing matrix
and a predefined dictionary, the average-case performance can be further
improved by employing an optimized multidimensional sensing matrix
and a learned multilinear sparsifying dictionary. In this paper, we
propose a joint optimization approach of the sensing matrix and dictionary
for a TCS system. For the sensing matrix design in TCS, an extended
separable approach with a closed form solution and a novel iterative
non-separable method are proposed when the multilinear dictionary
is fixed. In addition, a multidimensional dictionary learning method
that takes advantages of the multidimensional structure is derived,
and the influence of sensing matrices is taken into account in the
learning process. A joint optimization is achieved via alternately
iterating the optimization of the sensing matrix and dictionary. Numerical
experiments using both synthetic data and real images demonstrate
the superiority of the proposed approaches.\end{abstract}


\section{Introduction}

The traditional signal acquisition-and-compression paradigm removes
the signal redundancy and preserves the essential contents of signals
to achieve savings on storage and transmission, where the minimum
sampling ratio is restricted by the Shannon-Nyquist Theorem at the
signal sampling stage. The wasteful process of sensing-then-compressing
is replaced by directly acquiring the compressed version of signals
in Compressive Sensing (CS) \cite{Candes2006,Donoho2006,Candes2008},
a new sampling paradigm that leverages the fact that most signals
have sparse representations (i.e., there are only a few non-zero coefficients)
in some suitable basis. Successful reconstruction of such signals
is guaranteed for a sufficient number of randomly taken samples that
are far fewer in number than that required in the Shannon-Nyquist
Theorem. Therefore CS is very attractive for applications such as
medical imaging and wireless sensor networks where data acquisition
is expensive \cite{Lustig2007,Chen2011(2)}.

Achieving successful CS reconstruction has been characterized by a
number of properties, e.g., the Restricted Isometry Property (RIP)
\cite{Candes2006}, the mutual coherence \cite{donoho2006stable}
and the null space property \cite{Donoho2006}. These properties have
been used to provide sufficient conditions on sensing matrices and
to quantify the worst-case reconstruction performance \cite{Candes2008a,donoho2006stable,Donoho2006}.
Random matrices such as Gaussian or Bernoulli matrices have been shown
to fulfill these conditions, and hence are widely used as the sensing
matrix in CS applications. In view of the fact that the mainstream
view in the signal processing community considers the average-case
performance rather than the worst-case performance, later on, it is
shown that the average-case reconstruction performance can be further
enhanced by optimizing the sensing matrix according to the aforementioned
conditions, e.g., \cite{Elad2007,Xu2010,Chen2013,Li2013,Cleju2014}.
On the other hand, instead of using a fixed signal-sparsifying basis,
e.g., a Discrete Wavelet Transform (DWT), one can further enhance
CS performance by employing a basis which is learned from a training
data set to abstract the basic atoms that compose the signal ensemble.
The process of learning such a basis is referred to as ``sparsifying
dictionary learning'' and it has been widely investigated in the
literature \cite{Engan2000,Aharon2006,Tovsic2011,Sahoo2013,Dai2012}.
In addition, by further exploiting the interaction between the sensing
matrix and the sparsifying dictionary, joint optimization of the two
has also been considered in \cite{Duarte2009,Chen2013dictionary,Bai2015}.

However, in the process of sensing and reconstruction, the conventional
CS framework considers vectorized signals, and multidimensional signals
are mapped in a vector format in a CS system. At the sensing node,
such a vectorization requires the hardware to be capable of simultaneously
multiplexing along all data dimensions, which is hard to achieve especially
when one of the dimensions is along a timeline. Secondly, a real-world
vectorized signal requires an enormous sensing matrix that has as
many columns as the number of signal elements. Consequently such an
approach imposes large demands on the storage and processing power.
In addition, the vectorization also results in a loss of structure
along the various dimensions, the presence of which is beneficial
for developing efficient reconstruction algorithms. For these reasons,
applying conventional CS to applications that involve multidimensional
signals is challenging.

Extending CS to multidimensional signals has attracted growing interests
over the past few years. Most of the related work in the literature
focuses on CS for 2D signals (i.e., matrices), e.g., matrix completion
\cite{Recht2010,Candes2012:matrixCom}, and the reconstruction of
sparse and low rank matrices \cite{Golbabaee2012,Chartrand2012,Otazo2015}.
In \cite{Duarte2012}, Kronecker product matrices are proposed for
use in CS systems, which makes it possible to partition the sensing
process along signal dimensions and paves the way to developing CS
for tensors, i.e., signals with two or more dimensions. Tensor CS
(TCS) has been studied in \cite{Sidiropoulos2012,Friendland2014,Caiafa2013,Caiafa:2013},
where the main focus is on algorithm development for reconstruction.
To the best of our knowledge, there is no prior work concerning the
enhancement of TCS via optimizing the sensing matrices at various
dimensions in a tensor. In addition, although dictionary learning
techniques have been considered for tensors \cite{Seibert2014,Roemer2014,Peng2014},
it is still not clear how to conduct tensor dictionary learning to
incorporate the influence of sensing matrices in TCS.

In this paper, we investigate joint sensing matrix design and dictionary
learning for TCS systems. Unlike the optimization for a conventional
CS system where a single sensing matrix and a sparsifying basis for
vectorized signals are obtained, we produce a multiplicity of them
functioning along various tensor dimensions, thereby maintaining the
advantages of TCS. The contributions of this work are as follows: 
\begin{itemize}
\item We are the first to consider the optimization of a multidimensional
sensing matrix and dictionary for a TCS system and a joint optimization
of the two is designed, which also includes particular cases of optimizing
the sensing matrix for a given multilinear dictionary and learning
the dictionary for a given multidimensional sensing matrix.
\item We propose a separable approach for sensing matrix design by extending
the existing work for conventional CS. In this approach, the optimization
is proved to be separable, i.e., the sensing matrix along each dimension
can be independently optimized, and the approach has closed form solution.
\item We put forth a non-separable method for sensing matrix design using
a combination of the state-of-art measures for sensing matrix optimization.
This approach leads to the best reconstruction performance in our
comparison, but it is iterative and hence needs more computing power
to implement.
\item We propose a multidimensional dictionary learning approach that couples
the optimization of the multidimensional sensing matrix. This approach
extends KSVD \cite{Aharon2006} and coupled-KSVD \cite{Duarte2009}
to take full advantages of the multidimensional structure in tensors
with a reduced number of iterations required for the update of dictionary
atoms. 
\end{itemize}
The proposed approaches are demonstrated to enhance the performance
of existing TCS systems via the use of extensive simulations using
both synthetic data and real images. 

The remainder of this paper is organized as follows. Section \ref{sec:CS and TCS}
formulates CS and TCS, and introduces the related theory. Section
\ref{sec:PhiDesign} reviews the sensing matrix design approaches
for CS and presents the proposed methods for TCS sensing matrix design.
In Section \ref{sec:PsiDesign}, the related dictionary learning techniques
are reviewed, followed by the elaboration of the proposed multidimensional
dictionary learning approach and the joint optimization algorithm
is presented. Experimental results are given in Section \ref{sec:simulations}
and Section \ref{sec:conclusions} concludes the paper.

\subsection{Multilinear Algebra and Notations}

Boldface lower-case letters, boldface upper-case letters and non-boldface
letters denote vectors, matrices and scalars, respectively. A mode-$n$
tensor is an $n$-dimensional array $\underline{\mathbf{X}}\in\mathbb{R}^{N_{1}\times...\times N_{n}}$.
The mode-$i$ vectors of a tensor are determined by fixing every index
except the one in the mode $i$ and the slices of a tensor are its
two dimensional sections determined by fixing all but two indices.
By arranging all the mode-$i$ vectors as columns of a matrix, the
mode-$i$ unfolding matrix $\mathbf{X}_{(i)}\in\mathbb{R}^{N_{i}\times N_{1}...N_{i-1}N_{i+1}...N_{n}}$
is obtained. The mode-$k$ tensor by matrix product is defined as:
$\underline{\mathbf{Z}}=\underline{\mathbf{X}}\times_{k}\mathbf{A},$
where $\mathbf{A}\in\mathbb{R}^{J\times N_{k}}$, $\underline{\mathbf{Z}}\in\mathbb{R}^{N_{1}\times...\times N_{k-1}\times J\times N_{k+1}\times...\times N_{n}}$
and it is calculated by: $\underline{\mathbf{Z}}=fold_{i}(\mathbf{A}\mathbf{X}_{(i)})$,
where $fold_{i}(\cdot)$ means folding up a matrix along mode $i$
to a tensor. The matrix Kronecker product and vector outer product
are denoted by $\mathbf{A}\otimes\mathbf{B}$ and $\mathbf{a}\circ\mathbf{b}$,
respectively. The $l_{p}$ norm of a vector is defined as: $||\mathbf{x}||_{p}=(\sum_{\mathit{i}=1}^{\mathit{n}}|\mathit{x_{i}}|^{\mathit{p}})^{\frac{1}{\mathit{p}}}$.
For vectors, matrices and tensors, the $l_{0}$ norm is given by the
number of nonzero entries. $\mathbf{I}_{N}$ denotes the $N\times N$
identity matrix. The operator $(\cdot)^{-1}$, $(\cdot)^{T}$ and
$tr(\cdot)$ represent matrix inverse, matrix transpose and the trace
of a matrix, respectively. The number of elements for a vector, matrix
or tensor is denoted by $len(\cdot)$.

\section{Compressive Sensing (CS) and Tensor Compressive Sensing (TCS)}

\label{sec:CS and TCS}

\subsection{Sensing Model}

Consider a multidimensional signal $\underline{\mathbf{X}}\in\mathbb{R}^{N_{1}\times...\times N_{n}}$.
Conventional CS takes measurements from its vectorized version via:
\begin{equation}
\mathbf{y}=\boldsymbol{\Phi}\mathbf{x}+\mathbf{e},\label{eq:CS sensing}
\end{equation}
where $\mathbf{x}\in\mathbb{R}^{N}\ (N=\prod_{i}N_{i})$ denotes the
vectorized signal, $\boldsymbol{\Phi}\in\mathbb{R}^{M\times N}\ (M<N)$
is the sensing matrix, $\mathbf{y}\in\mathbb{R}^{M}$ represents the
measurement vector and $\mathbf{e}\in\mathbb{R}^{M}$ is a noise term.
The vectorized signal is assumed to be sparse in some sparsifying
basis $\boldsymbol{\Psi}\in\mathbb{R}^{N\times\hat{N}}\ (N\leq\hat{N})$,
i.e., 
\begin{equation}
\mathbf{x}=\boldsymbol{\Psi}\mathbf{s},
\end{equation}
where $\mathbf{s}\in\mathbb{R}^{\hat{N}}$ is the sparse representation
of $\mathbf{x}$ and it has only $K\ (K\ll\hat{N})$ non-zero coefficients.
Thus the sensing model can be rewritten as:
\begin{equation}
\mathbf{y}=\boldsymbol{\Phi}\boldsymbol{\Psi}\mathbf{s}+\mathbf{e}=\mathbf{A}\mathbf{s}+\mathbf{e},\label{eq:CS eqSensing}
\end{equation}
where $\mathbf{A}=\boldsymbol{\Phi\Psi}\in\mathbb{R}^{M\times\hat{N}}$
is the equivalent sensing matrix.

Even though CS has been successfully applied to practical sensing
systems \cite{Duarte2008,Marcia2009,Majidzadeh2010}, the sensing
model has a few drawbacks when it comes to tensor signals. First of
all, the multidimensional structure presented in the original signal
$\underline{\mathbf{X}}$ is omitted due to the vectorization, which
loses information that can lead to efficient reconstruction algorithms.
Besides, as stated by (\ref{eq:CS sensing}), the sensing system is
required to operate along all dimensions of the signal simultaneously,
which is difficult to achieve in practice. Furthermore, the size of
$\boldsymbol{\Phi}$ associated with the vectorized signal becomes
too large to be practical for applications involving multidimensional
signals. 

TCS tackles these problems by utilizing separable sensing operators
along tensor modes and its sensing model is: 
\begin{equation}
\underline{\mathbf{Y}}=\underline{\mathbf{X}}\times_{1}\boldsymbol{\Phi}_{1}\times_{2}\boldsymbol{\Phi}_{2}...\times_{n}\boldsymbol{\Phi}_{n}+\underline{\mathbf{E}},\label{eq:TCS sensing}
\end{equation}
where $\underline{\mathbf{Y}}\in\mathbb{R}^{M_{1}\times...M_{n}}$
represents the measurement, $\underline{\mathbf{E}}\in\mathbb{R}^{M_{1}\times...M_{n}}$
denotes the noise term, $\boldsymbol{\Phi}_{i}\in\mathbb{R}^{M_{i}\times N_{i}}\ (i=1,...,n)$
are sensing matrices and $M_{i}<N_{i}$. The multidimensional signal
is assumed to be sparse in a separable sparsifying basis $\boldsymbol{\Psi}_{i}\in\mathbb{R}^{N_{i}\times\hat{N}_{i}}\ (i=1,...,n)$,
i.e.,
\begin{equation}
\underline{\mathbf{X}}=\underline{\mathbf{S}}\times_{1}\boldsymbol{\Psi}_{1}\times_{2}\boldsymbol{\Psi}_{2}...\times_{n}\boldsymbol{\Psi}_{n},
\end{equation}
where $\underline{\mathbf{S}}\in\mathbb{R}^{\hat{N}_{1}\times...\hat{N}_{n}}$
is the sparse representation that has only $K\ (K\ll\prod_{i}\hat{N}_{i})$
non-zero coefficients. The equivalent sensing model can then be written
as: 
\begin{equation}
\underline{\mathbf{Y}}=\underline{\mathbf{S}}\times_{1}\mathbf{A}_{1}\times_{2}\mathbf{A}_{2}...\times_{n}\mathbf{A}_{n},\label{eq:TCS eqSensing}
\end{equation}
where $\mathbf{A}_{i}=\boldsymbol{\Phi}_{i}\boldsymbol{\Psi}_{i}\ (i=1,...,n)$
are the equivalent sensing matrices.

Using the TCS sensing model in (\ref{eq:TCS sensing}), the sensing
procedure in (\ref{eq:CS sensing}) is partitioned into a few processes
having smaller sensing matrices $\boldsymbol{\Phi}_{i}\in\mathbb{R}^{M_{i}\times N_{i}}\ (i=1,...,n)$
and yet it maintains the multidimensional structure of the original
signal $\underline{\mathbf{X}}$. It is also useful to mention that
the TCS model in (\ref{eq:TCS eqSensing}) is equivalent to:
\begin{equation}
\mathbf{y}=(\mathbf{A}_{n}\otimes\mathbf{A}_{n-1}\otimes...\otimes\mathbf{A}_{1})\mathbf{s},\label{eq:CSvsTCS}
\end{equation}
as derived in \cite{Caiafa2013}. By denoting $\overline{\mathbf{A}}=\mathbf{A}_{n}\otimes\mathbf{A}_{n-1}\otimes...\otimes\mathbf{A}_{1}$,
it becomes a conventional CS model akin to (\ref{eq:CS eqSensing}),
except that the sensing matrix in (\ref{eq:CSvsTCS}) has a multilinear
structure.

\subsection{Signal Reconstruction }

\label{sub:SigReconModel}

In conventional CS, the problem of reconstructing $\mathbf{s}$ from
the measurement vector $\mathbf{y}$ captured using (\ref{eq:CS eqSensing})
is modeled as a $l_{0}$ minimization problem as follows: 
\begin{equation}
\underset{\mathbf{s}}{min}\ ||\mathbf{s}||_{0},\ s.t.\ ||\mathbf{y}-\mathbf{As}||\leq\varepsilon,\label{eq:CS_l0_min}
\end{equation}
where $\varepsilon$ is a tolerance parameter. Many algorithms have
been developed to solve this problem, including Basis Pursuit (BP)
\cite{Candes2005,Candes2006,Candes2008,Donoho2006}, i.e., conducting
convex optimization by relaxing the $l_{0}$ norm in (\ref{eq:CS_l0_min})
as the $l_{1}$ norm, and greedy algorithms such as Orthogonal Matching
Pursuit (OMP) \cite{Tropp2007} and Iterative Hard Thresholding (IHT)
\cite{Blumensath2009}. The reconstruction performance of the $l_{1}$
minimization approach has been studied in \cite{Candes2005,Candes2008a},
where the well known Restricted Isometry Property (RIP) was introduced
to provide a sufficient condition for successful signal recovery.

\textit{\textcolor{black}{Definition 1: }}\textcolor{black}{A matrix
$\mathbf{A}$ satisfies the RIP of order $K$ with a the Restricted
Isometry Constant (RIC) ${\color{black}\delta}_{{\color{black}K}}$
being the smallest number such that
\begin{equation}
(1-\delta_{K})||\mathbf{s||_{\mathrm{2}}^{\mathrm{2}}}\leq||\mathbf{As||_{\mathrm{2}}^{\mathrm{2}}}\leq(1+\delta_{K})||\mathbf{s||_{\mathrm{2}}^{\mathrm{2}}}\label{eq:RIP}
\end{equation}
holds for all} \textcolor{black}{$\mathbf{s}$ with $||\mathbf{s}||_{0}\leq K$.
}\hfill{}$\blacksquare$

\textit{\textcolor{black}{Theorem 1:}}\textcolor{black}{{} Assume that
${\color{black}\delta_{2K}<\sqrt{2}-1}$ and $||\mathbf{e}||_{2}\leq\varepsilon$.
Then the solution ${\color{black}\mathbf{\hat{s}}}$ to (\ref{eq:CS_l0_min})
obeys
\begin{equation}
||{\color{black}\mathbf{\hat{s}-s\mathrm{||_{2}\leq\mathit{C_{0}}\mathbf{\mathit{K}\mathrm{^{-1/2}}}||\mathbf{s}}-\mathrm{\mathbf{s}}_{\mathit{K}}}}||_{1}+\mathit{C_{1}}\varepsilon\label{eq:error bound}
\end{equation}
where ${\color{black}\mathit{C_{0}=\frac{\mathrm{2+(2\sqrt{2}-2)\delta_{2\mathit{K}}}}{\mathrm{1-(\sqrt{2}+1)\delta_{2\mathit{K}}}}}}$,
${\color{black}\mathit{C_{1}=\frac{4\sqrt{\mathrm{1+\delta_{2\mathit{K}}}}}{\mathrm{1-(\sqrt{2}+1)\delta_{2\mathit{K}}}}}}$,
${\color{black}\delta_{2\mathit{K}}}$ is the RIC of matrix ${\color{black}\mathbf{A}}$,
$\mathbf{s}_{K}$ is an approximation of ${\color{black}\mathbf{s}}$
with all but the ${\color{black}\mathit{K}}$ largest entries set
to zero.}\hfill{}$\blacksquare$

The previous theorem states that for the noiseless case, any sparse
signal with fewer than $K$ non-zero coefficients can be exactly recovered
if the RIC of the equivalent sensing matrix satisfies \textcolor{black}{${\color{black}\delta_{2K}<\sqrt{2}-1}$;
while for the noisy case and the not exactly sparse case, the reconstructed
signal is still a good approximation of the original signal under
the same condition. The theoretical guarantees of successful reconstruction
for the greedy approaches have also been investigated in \cite{Tropp2007,Blumensath2009}. }

\textcolor{black}{The RIP essentially measures the quality of the
equivalent sensing matrix $\mathbf{A}$, which closely relates to
the design of $\boldsymbol{\Phi}$ and $\boldsymbol{\Psi}$. However,
since the RIP is not tractable, another measure is often used for
CS projection design, i.e., the mutual coherence of $\mathbf{A}$
\cite{donoho2006stable} and it is defined by: 
\begin{equation}
\mu(\mathbf{A})=\underset{1\leq i,\ j\leq\hat{N},\ i\neq j}{max}|\mathbf{a}_{i}^{T}\mathbf{a}_{j}|,
\end{equation}
}where $\mathbf{a}_{i}$ denotes the $i$th column of $\mathbf{A}$.
It has been shown that the reconstruction error of the $l_{1}$ minimization
problem is bounded if $\mu(\mathbf{A})<1/(4K-1)$. Based on the concept
of mutual coherence, optimal projection design approaches are derived,
e.g., in \cite{Elad2007,Duarte2009,Xu2010}.

When it comes to TCS, the reconstruction approaches for CS can still
be utilized owing to the relationship in (\ref{eq:CSvsTCS}). However,
for the algorithms where explicit usage of $\overline{\mathbf{A}}$
is required, e.g, OMP, the implementation is restricted by the large
dimension of $\overline{\mathbf{A}}$. By extending the CS reconstruction
approaches to utilize tensor-based operations, TCS reconstruction
algorithms employing only small matrices $\mathbf{A}_{i}\ (i=1,...,n)$
have been developed in \cite{Caiafa2013,Caiafa:2013,Rivenson2009a,Rivenson2009b}.
These methods maintain the theoretical guarantees of conventional
CS when $\overline{\mathbf{A}}$ obeys the condition on the RIC or
the mutual coherence, but reduce the computational complexity and
relax the storage memory requirement.

Even so, the conditions on $\overline{\mathbf{A}}$ are not intuitive
for a practical TCS system, which explicitly utilizes multiple separable
sensing matrices $\mathbf{A}_{i}\ (i=1,...,n)$ instead of a single
matrix $\overline{\mathbf{A}}$. Fortunately, the authors of \cite{Duarte2012}
have derived the following relationships to clarify the corresponding
conditions on $\mathbf{A}_{i}\ (i=1,...,n)$.

\textit{\textcolor{black}{Theorem 2: }}\textcolor{black}{Let }$\mathbf{A}_{i}\ (i=1,...,n)$
be matrices with RICs $\delta_{K}(\mathbf{A}_{1})$, ..., $\delta_{K}(\mathbf{A}_{n})$,
respectively, and their mutual coherence are $\mu(\mathbf{A}_{1})$,
..., $\mu(\mathbf{A}_{n})$. Then for the matrix $\overline{\mathbf{A}}=\mathbf{A}_{n}\otimes\mathbf{A}_{n-1}\otimes...\otimes\mathbf{A}_{1}$,
we have 
\begin{align}
\mu(\overline{\mathbf{A}}) & =\prod_{i=1}^{n}\mu(\mathbf{A}_{i}),\\
\delta_{K}(\overline{\mathbf{A}}) & \leq\prod_{i=1}^{n}(1+\delta_{K}(\mathbf{A}_{i}))-1.
\end{align}
 \hfill{}$\blacksquare$

In \cite{Duarte2012}, these relationships are then utilized to derive
the reconstruction error bounds for a TCS system.

\section{Optimized Multilinear Projections for TCS}

\label{sec:PhiDesign}

In this section, we show how to optimize the multilinear sensing matrix
when the dictionaries $\boldsymbol{\Psi}_{i}\ (i=1,...,n)$ for each
dimension are fixed. We first introduce the related design approaches
for CS, then present the proposed methods for TCS, including a separable
and a non-separable design approach.

\subsection{Sensing Matrix Design for CS }

\label{sub:CS sensing design}

We observe that the sufficient conditions on the RIC or the mutual
coherence for successful CS reconstruction, as reviewed in Section
\ref{sub:SigReconModel}, only describe the worst case bound, which
means that the average recovery performance is not reflected. In fact,
the most challenging part of CS sensing matrix design lies in deriving
a measure that can directly reveal the expected-case reconstruction
accuracy. 

In \cite{Elad2007}, Elad \textit{et al.} proposed the notion of averaged
mutual coherence, based on which an iterative algorithm is derived
for optimal sensing matrix design. This approach aims to minimize
the largest absolute values of the off-diagonal entries in the Gram
matrix of $\mathbf{A}$, i.e., $\mathbf{G}_{\mathbf{A}}=\mathbf{A}^{T}\mathbf{A}$.
It has been shown to outperform a random Gaussian sensing matrix in
terms of reconstruction accuracy, but is time-consuming to construct
and can ruin the worst case guarantees by inducing large off-diagonal
values that are not in the original Gram matrix. In order to make
any subset of columns in $\mathbf{A}$ as orthogonal as possible,
Sapiro\textit{ et al.} proposed in \cite{Duarte2009} to make $\mathbf{G}_{\mathbf{A}}$
as close as possible to an identity matrix, i.e., $\boldsymbol{\Psi}^{T}\boldsymbol{\Phi}^{T}\boldsymbol{\Phi}\boldsymbol{\Psi}\approx\mathbf{I}_{\hat{N}}$.
It is then approximated by minimizing $||\boldsymbol{\Lambda}-\boldsymbol{\Lambda\Gamma}^{T}\boldsymbol{\Gamma\Lambda}||_{F}^{2}$,
where $\boldsymbol{\Gamma}$ comes from the eigen-decomposition of
$\boldsymbol{\Psi}^{T}\boldsymbol{\Psi}$, i.e., $\boldsymbol{\Psi}^{T}\boldsymbol{\Psi}=\mathbf{V}\boldsymbol{\Lambda}\mathbf{V}^{T}$,
and $\boldsymbol{\Gamma}=\boldsymbol{\Phi}\mathbf{V}$. This approach
is also iterative, but outperforms Elad's method. Considering the
fact that $\mathbf{A}$ has minimum coherence when the magnitudes
of all the off-diagonal entries of $\mathbf{G}_{\mathbf{A}}$ are
equal, Xu \textit{et al.} proposed an Equiangular Tight Frame (ETF)
based method in \cite{Xu2010}. The problem is modeled as: $min_{\mathbf{G}_{t}\in\mathcal{H}}||\boldsymbol{\Psi}^{T}\boldsymbol{\Phi}^{T}\boldsymbol{\Phi}\boldsymbol{\Psi}-\mathbf{G}_{t}||_{F}^{2}$,
where $\mathbf{G}_{t}$ is the target Gram matrix and $\mathcal{H}$
is the set of the ETF Gram matrices. Improved performance has been
observed for the obtained sensing matrix.

More recently, based on the same idea as Sapiro, the problem of 
\begin{equation}
\underset{\boldsymbol{\Phi}}{min}\ ||\mathbf{I}_{\hat{N}}-\boldsymbol{\Psi}^{T}\boldsymbol{\Phi}^{T}\boldsymbol{\Phi}\boldsymbol{\Psi}||_{F}^{2}\label{eq: Li's design}
\end{equation}
has been considered and an analytical solution has been derived in
\cite{Li2013}. Meanwhile, in \cite{Chen2013,Chen2012}, it has been
shown that in order to achieve good expected-case Mean Squared Error
(MSE) performance, the equivalent sensing matrix ought to be close
to a Parseval tight frame, thus leading to the following design approach:
\begin{equation}
\underset{\boldsymbol{\Phi}}{min}\ ||\boldsymbol{\Phi}||_{F}^{2},\ s.t.\ \boldsymbol{\Phi}\boldsymbol{\Psi}\boldsymbol{\Psi}^{T}\boldsymbol{\Phi}^{T}=\mathbf{I}_{M},\label{eq:Chen's design}
\end{equation}
where $||\boldsymbol{\Phi}||_{F}^{2}$ is the sensing cost that also
affects the reconstruction accuracy (as verified in \cite{Chen2013,Chen2012}).
A closed form solution to this problem was also obtained in \cite{Chen2013,Chen2012}.
These approaches have further improved the average reconstruction
performance for a CS system that is able to employ the optimized sensing
matrix.

On the other hand, using the model of Xu's method \cite{Xu2010},
Cleju \cite{Cleju2014} proposed to take $\mathbf{G}_{t}=\boldsymbol{\Psi}^{T}\boldsymbol{\Psi}$
so that the equivalent sensing matrix has similar properties to those
of $\boldsymbol{\Psi}$; and Bai \textit{et al. }\cite{Bai2015} proposed
combining the ETF Grams and that proposed by Cleju to solve: $min_{\mathbf{G}_{t}\in\mathcal{H}}(1-\beta)||\boldsymbol{\Psi}^{T}\boldsymbol{\Psi}-\boldsymbol{\Psi}^{T}\boldsymbol{\Phi}^{T}\boldsymbol{\Phi}\boldsymbol{\Psi}||_{F}^{2}+\beta||\mathbf{G}_{t}-\boldsymbol{\Psi}^{T}\boldsymbol{\Phi}^{T}\boldsymbol{\Phi}\boldsymbol{\Psi}||_{F}^{2}$,
where $\beta$ is a trade-off parameter. Promising results of these
methods are demonstrated.

\subsection{Multidimensional Sensing Matrix Design for TCS}

\label{sub:SensingDesign}

In contrast to the aforementioned methods, we consider optimization
of the sensing matrix for TCS. Compared to the design process in conventional
CS, the main distinction for the TCS is that we would like to optimize
multiple separable sensing matrices $\boldsymbol{\Phi}_{i}\ (i=1,...,n)$,
rather than a single matrix $\boldsymbol{\Phi}$. In this section,
in addition to extending the approaches in (\ref{eq: Li's design})
and (\ref{eq:Chen's design}) to the TCS case, we also propose a new
approach for TCS sensing matrices design by combining the state-of-art
ideas in \cite{Chen2013,Cleju2014,Bai2015}. To simplify our exposition,
we elaborate our methods in the following sections for the case of
$n=2$, i.e., the tensor signal becomes a matrix, but note that the
methods can be straightforwardly extended to an $n$ mode tensor case
($n>2$).

As reviewed in Section \ref{sub:SigReconModel}, the performance of
existing TCS reconstruction algorithms relies on the quality of $\overline{\mathbf{A}}$,
where $\overline{\mathbf{A}}=\mathbf{A}_{2}\otimes\mathbf{A}_{1}$
when $n=2$. Therefore, when the multilinear dictionary $\overline{\boldsymbol{\Psi}}=\boldsymbol{\Psi}_{2}\otimes\boldsymbol{\Psi}_{1}$
is given, one can optimize $\overline{\boldsymbol{\Phi}}$ (where
$\overline{\boldsymbol{\Phi}}=\boldsymbol{\Phi}_{2}\otimes\boldsymbol{\Phi}_{1}$)
using the methods for CS as introduced in Section \ref{sub:CS sensing design}. 

However, when implementing a TCS system, it is still necessary to
obtain the separable matrices, i.e., $\boldsymbol{\Phi}_{1}$ and
$\boldsymbol{\Phi}_{2}$. One intuitive solution is to design $\overline{\boldsymbol{\Phi}}$
using the aforementioned approaches for CS and then to decompose $\overline{\boldsymbol{\Phi}}$
by solving the following problem:
\begin{equation}
\underset{\boldsymbol{\Phi}_{1},\boldsymbol{\Phi}_{2}}{min}\ ||\overline{\boldsymbol{\Phi}}-\boldsymbol{\Phi}_{2}\otimes\boldsymbol{\Phi}_{1}||_{F}^{2},\label{eq:NKP}
\end{equation}
which has been studied as a Nearest Kronecker Product (NKP) problem
in \cite{Van2000}. But this is not a feasible solution for TCS sensing
matrix design. First of all, $\overline{\boldsymbol{\Phi}}$ can only
be exactly decomposed as $\boldsymbol{\Phi}_{2}\otimes\boldsymbol{\Phi}_{1}$
when a certain permutation of $\overline{\boldsymbol{\Phi}}$ has
rank 1 \cite{Van2000}, which is not the case for most sensing strategies.
When the term in (\ref{eq:NKP}) is minimized to a non-zero value,
the solution $\hat{\boldsymbol{\Phi}}_{1},\ \hat{\boldsymbol{\Phi}}_{2}$
leads to a sensing matrix $\hat{\boldsymbol{\Phi}}_{2}\otimes\hat{\boldsymbol{\Phi}}_{1}$,
which may not satisfy the condition of the sensing matrix $\overline{\boldsymbol{\Phi}}$
for good CS recovery (e.g., the requirement on the mutual coherence),
thereby ruining the reconstruction guarantees. Secondly, to solve
(\ref{eq:NKP}), explicit storage of $\overline{\boldsymbol{\Phi}}$
is necessary, which is restrictive for high dimensional problems.
In addition, when the number of tensor modes increases, the problem
becomes more complex to solve.

Therefore, we aim to optimize $\boldsymbol{\Phi}_{1}$ and $\boldsymbol{\Phi}_{2}$
directly without knowing $\overline{\boldsymbol{\Phi}}$. Extending
(\ref{eq: Li's design}) and (\ref{eq:Chen's design}), we first propose
a method that is shown to be separable as independent sub-design-problems.
Then a non-separable design approach is presented and a gradient based
algorithm is derived.

\subsubsection{A Separable Design Approach}

The proposed separable design approach (Approach I) is as follows:
\begin{equation}
\underset{\boldsymbol{\Phi}_{1},\boldsymbol{\Phi}_{2}}{min}\ ||\mathbf{I}_{\hat{N}_{1}\hat{N}_{2}}-(\boldsymbol{\Psi}_{2}^{T}\otimes\boldsymbol{\Psi}_{1}^{T})(\boldsymbol{\Phi}_{2}^{T}\otimes\boldsymbol{\Phi}_{1}^{T})(\boldsymbol{\Phi}_{2}\otimes\boldsymbol{\Phi}_{1})(\boldsymbol{\Psi}_{2}\otimes\boldsymbol{\Psi}_{1})||_{F}^{2},\label{eq:Approach I}
\end{equation}
and it is an extension of (\ref{eq: Li's design}) to the case when
a multilinear sensing matrix is employed. The solution of (\ref{eq:Approach I})
is presented in Theorem 3 and Approach I is also summarized in Algorithm
1.

\textit{Theorem 3:} Assume for $i=1,\ 2,$ $\bar{N}_{i}=rank(\boldsymbol{\Psi}_{i})$,
$\boldsymbol{\Psi}_{i}=\mathbf{U}_{\boldsymbol{\Psi}_{i}}\left[\begin{array}{cc}
\boldsymbol{\Lambda}_{\boldsymbol{\Psi}_{i}} & \mathbf{0}\\
\mathbf{0} & \mathbf{0}
\end{array}\right]\mathbf{V}_{\boldsymbol{\Psi}_{i}}^{T}$ is an SVD of $\boldsymbol{\Psi}_{i}$ and $\boldsymbol{\Lambda}_{\boldsymbol{\Psi}_{i}}\in\mathbb{R}^{\bar{N}_{i}\times\bar{N}_{i}}$.
Let $\hat{\boldsymbol{\Phi}}_{i}\in\mathbb{R}^{M_{i}\times N_{i}}\ (i=1,\ 2)$
be matrices with $rank(\hat{\boldsymbol{\Phi}}_{i})=M_{i}$ and $M_{i}\leq\bar{N}_{i}$
is assumed. Then 
\begin{itemize}
\item the following equation is a solution to (\ref{eq:Approach I}): 
\begin{equation}
\hat{\boldsymbol{\Phi}}_{i}=\mathbf{U}\left[\begin{array}{cc}
\mathbf{I}_{M_{i}} & \mathbf{0}\end{array}\right]\left[\begin{array}{cc}
\mathbf{V}^{T}\boldsymbol{\Lambda}_{\boldsymbol{\Psi}_{i}}^{-1} & \mathbf{0}\\
\mathbf{0} & \mathbf{0}
\end{array}\right]\mathbf{U}_{\boldsymbol{\Psi}_{i}}^{T},\label{eq:solutionApproI}
\end{equation}
where $i=1,\ 2$, $\mathbf{U}\in\mathbb{R}^{M_{i}\times M_{i}}$ and
$\mathbf{V}\in\mathbb{R}^{\bar{N}_{i}\times\bar{N}_{i}}$ are arbitrary
orthonormal matrices; 
\item the resulting equivalent sensing matrices $\hat{\mathbf{A}}_{i}=\hat{\boldsymbol{\Phi}}_{i}\boldsymbol{\Psi}_{i}\ (i=1,\ 2)$
are Parseval tight frames, i.e., $||\hat{\mathbf{A}}_{i}^{T}\mathbf{z}||_{2}=||\mathbf{z}||_{2}$,
where $\mathbf{z}\in\mathbb{R}^{\hat{N}_{i}}$ is an arbitrary vector.
\item the minimum of (\ref{eq:Approach I}) is $\hat{N}_{1}\hat{N}_{2}-M_{1}M_{2}$; 
\item separately solving the sub-problems
\begin{equation}
\underset{\boldsymbol{\Phi}_{i}}{min}\ ||\mathbf{I}_{\hat{N}_{i}}-\boldsymbol{\Psi}_{i}^{T}\boldsymbol{\Phi}_{i}^{T}\boldsymbol{\Phi}_{i}\boldsymbol{\Psi}_{i}||_{F}^{2}\label{eq:sub_prob_I}
\end{equation}
for $i=1,\ 2$ leads to the same solutions as (\ref{eq:solutionApproI})
and the resulting objective in (\ref{eq:Approach I}) has the same
minimum, i.e., $\hat{N}_{1}\hat{N}_{2}-M_{1}M_{2}$.\hfill{}$\blacksquare$
\end{itemize}
\textit{Proof:} The proof is given in Appendix \ref{sec:AppendixI}.

\begin{algorithm}[h]
\caption{\textbf{Design Approach I}}

\textbf{Input: }$\boldsymbol{\Psi}_{i}$ $(i=1,\ 2)$. 

\textbf{Output:} $\hat{\boldsymbol{\Phi}}_{i}$ $(i=1,\ 2)$.

1:\textbf{ for} $i=1,\ 2$ \textbf{do}

2: \hspace{1em}Calculate optimized $\hat{\boldsymbol{\Phi}}_{i}$
using (\ref{eq:solutionApproI});

3: \textbf{end}

4:\textbf{ }Normalization for $i=1,\ 2$: $\hat{\boldsymbol{\Phi}}_{i}=\sqrt{N_{i}}\hat{\boldsymbol{\Phi}}_{i}/||\hat{\boldsymbol{\Phi}}_{i}||_{F}$.
\end{algorithm}

Clearly, Approach I is separable, which means that we can independently
design each $\boldsymbol{\Phi}_{i}$ according to the corresponding
sparsifying dictionary $\boldsymbol{\Psi}_{i}$ in mode $i$. This
observation stays consistent when we consider the situation in an
alternative way. Applying the method in (\ref{eq: Li's design}) to
acquire the optimal $\boldsymbol{\Phi}_{1}$ and $\boldsymbol{\Phi}_{2}$
independently, we are actually trying to make any subset of columns
in $\mathbf{A}_{1}$ and $\mathbf{A}_{2}$, respectively, as orthogonal
as possible. As a result, the matrix $\overline{\mathbf{A}}=\mathbf{A}_{2}\otimes\mathbf{A}_{1}$
that is obtained will also be as orthogonal as possible. This follows
from the fact that for any two columns of $\overline{\mathbf{A}}$,
we have 
\begin{align}
|\mathbf{\overline{a}}_{p}^{T}\mathbf{\overline{a}}_{q}| & =|[(\mathbf{a}_{2})_{l}^{T}\otimes(\mathbf{a}_{1})_{s}^{T}][(\mathbf{a}_{2})_{c}\otimes(\mathbf{a}_{1})_{d}]|\nonumber \\
 & =|[(\mathbf{a}_{2})_{l}^{T}(\mathbf{a}_{2})_{c}][(\mathbf{a}_{1})_{s}^{T}(\mathbf{a}_{1})_{d}]|,
\end{align}
where $\mathbf{\overline{a}}$, $\mathbf{a}_{1}$ and $\mathbf{a}_{2}$
denote the column of $\overline{\mathbf{A}}$, $\mathbf{A}_{1}$ and
$\mathbf{A}_{2}$, respectively, and $p,\ q,\ l,\ s,\ c,\ d$ are
the column indices. 

Using the second statement of Theorem 3, we can derive the following
corollary.

\textit{Corollary 1:} The solution in (\ref{eq:solutionApproI}) also
solves the following problems for $i=1,\ 2$: 
\begin{equation}
\underset{\boldsymbol{\Phi}_{i}}{min}\ ||\boldsymbol{\Phi}_{i}||_{F}^{2},\ s.t.\ \boldsymbol{\Phi}_{i}\boldsymbol{\Psi}_{i}\boldsymbol{\Psi}_{i}^{T}\boldsymbol{\Phi}_{i}^{T}=\mathbf{I}_{M_{i}},\label{eq:subII}
\end{equation}
which represent the separable sub-problems of the following design
approach:
\begin{gather}
\underset{\boldsymbol{\Phi}_{1},\boldsymbol{\Phi}_{2}}{min}\ ||\boldsymbol{\Phi}_{2}\otimes\boldsymbol{\Phi}_{1}||_{F}^{2},\label{eq:Approach II}\\
s.t.\ (\boldsymbol{\Phi}_{2}\otimes\boldsymbol{\Phi}_{1})(\boldsymbol{\Psi}_{2}\otimes\boldsymbol{\Psi}_{1})(\boldsymbol{\Psi}_{2}^{T}\otimes\boldsymbol{\Psi}_{1}^{T})(\boldsymbol{\Phi}_{2}^{T}\otimes\boldsymbol{\Phi}_{1}^{T})=\mathbf{I}_{M_{1}M_{2}},\nonumber 
\end{gather}
and it is in fact a multidimensional extension of the CS sensing matrix
design approach proposed in \cite{Chen2013}. \hfill{}$\blacksquare$

\textit{Proof:} Since the equivalent sensing matrices designed using
Approach I are Parseval tight frames, it follows from the derivation
in \cite{Chen2013} that the sub-problems in (\ref{eq:subII}) have
the same solution as in (\ref{eq:solutionApproI}). The problem in
(\ref{eq:Approach II}) can be proved separable simply by revealing
the fact that $||\boldsymbol{\Phi}_{2}\otimes\boldsymbol{\Phi}_{1}||_{F}^{2}=||\boldsymbol{\Phi}_{2}||_{F}^{2}||\boldsymbol{\Phi}_{1}||_{F}^{2}$,
and when $\boldsymbol{\Phi}_{i}\boldsymbol{\Psi}_{i}\boldsymbol{\Psi}_{i}^{T}\boldsymbol{\Phi}_{i}^{T}=\mathbf{I}_{M_{i}}$
is satisfied for both $i=1$ and 2, the constraint in (\ref{eq:Approach II})
is also satisfied. \hfill{}$\blacksquare$

By decomposing the original problems into independent sub-problems,
the sensing matrices can be designed in parallel and the problem becomes
easier to solve. However, the CS sensing matrix design approaches
are not always separable after being extended to the multidimensional
case, because a variety of different criteria can be used for sensing
matrix design as reviewed in Section \ref{sub:CS sensing design},
and in many cases the decomposition is not provable. We will propose
a non-separable approach in the following section.

\subsubsection{A Non-separable Design Approach}

Taking into account: i) the impact of sensing cost on reconstruction
performance \cite{Chen2013}; ii) the benefit of making the equivalent
sensing matrix so that it has similar properties to those of the sparsifying
dictionary \cite{Cleju2014}; and iii) the conventional requirement
on the mutual coherence, we put forth the following Design Approach
II: 
\begin{align}
\underset{\boldsymbol{\Phi}_{1},\boldsymbol{\Phi}_{2}}{min}\  & (1-\beta)||(\overline{\boldsymbol{\Psi}})^{T}\overline{\boldsymbol{\Psi}}-(\overline{\boldsymbol{\Psi}})^{T}(\overline{\boldsymbol{\Phi}})^{T}\overline{\boldsymbol{\Phi}}\overline{\boldsymbol{\Psi}}||_{F}^{2}\nonumber \\
 & +\alpha||\overline{\boldsymbol{\Phi}}||_{F}^{2}+\beta||\mathbf{I}_{\hat{N}_{1}\hat{N}_{2}}-(\overline{\boldsymbol{\Psi}})^{T}(\overline{\boldsymbol{\Phi}})^{T}\overline{\boldsymbol{\Phi}}\overline{\boldsymbol{\Psi}}||_{F}^{2},\label{eq:Approach III}
\end{align}
where $\overline{\boldsymbol{\Psi}}=\boldsymbol{\Psi}_{2}\otimes\boldsymbol{\Psi}_{1}$,
$\overline{\boldsymbol{\Phi}}=\boldsymbol{\Phi}_{2}\otimes\boldsymbol{\Phi}_{1}$,
$\alpha$ and $\beta$ are tuning parameters. As investigated in \cite{Chen2013}
and \cite{Bai2015}, $\alpha\geq0$ controls the sensing energy; while
$\beta\in[0,\ 1]$ balances the impact of the first and third terms
to achieve optimal performance under different conditions of the measurement
noise. The choice of these parameters will be investigated in Section
\ref{sub:Simulation_Phi}.

To solve (\ref{eq:Approach III}), we adopt a coordinate descent method.
Denoting the objective as $f(\boldsymbol{\Phi}_{1},\boldsymbol{\Phi}_{2})$,
we first compute its gradient with respect to $\boldsymbol{\Phi}_{1}$
and $\boldsymbol{\Phi}_{2}$, respectively, and the result is as follows:

\begin{align}
\frac{\partial f}{\partial\boldsymbol{\Phi}_{i}} & =4||\mathbf{G}_{\mathbf{A}_{j}}||_{F}^{2}(\mathbf{A}_{i}\mathbf{G}_{\mathbf{A}_{i}}\boldsymbol{\Psi}_{i}^{T})-4\beta||\mathbf{A}_{j}||_{F}^{2}(\mathbf{A}_{i}\boldsymbol{\Psi}_{i}^{T})\nonumber \\
 & +2\alpha||\boldsymbol{\Phi}_{j}||_{F}^{2}\boldsymbol{\Phi}_{i}+4(\beta-1)||\boldsymbol{\Psi}_{j}\mathbf{A}_{j}^{T}||_{F}^{2}(\mathbf{A}_{i}\mathbf{G}_{\boldsymbol{\Psi}_{i}}\boldsymbol{\Psi}_{i}^{T}),\label{eq:gradient_2D}
\end{align}
where $i,\ j\in\{1,\ 2\}$ and $j\neq i$, $\mathbf{G}_{\mathbf{A}_{i}}=\mathbf{A}_{i}^{T}\mathbf{A}_{i}$
and $\mathbf{G}_{\boldsymbol{\Psi}_{i}}=\boldsymbol{\Psi}_{i}^{T}\boldsymbol{\Psi}_{i}$. 

For generality, we also provide the result for the $n>2$ case as
follows: 
\begin{eqnarray}
\frac{\partial f}{\partial\boldsymbol{\Phi}_{i}} & = & 4\omega_{i}(\mathbf{A}_{i}\mathbf{G}_{\mathbf{A}_{i}}\boldsymbol{\Psi}_{i}^{T})-4\beta\theta_{i}(\mathbf{A}_{i}\boldsymbol{\Psi}_{i}^{T})\nonumber \\
 & + & 2\alpha\tau_{i}\boldsymbol{\Phi}_{i}+(4\beta-4)\rho_{i}(\mathbf{A}_{i}\mathbf{G}_{\boldsymbol{\Psi}_{i}}\boldsymbol{\Psi}_{i}^{T}),
\end{eqnarray}
where $i,\ j\in\{1,...,n\}$ and $j\neq i$, $\omega_{i}=\prod_{j}||\mathbf{G}_{\mathbf{A}_{j}}||_{F}^{2}$,
$\theta_{i}=\prod_{j}||\mathbf{A}_{j}||_{F}^{2}$, $\tau_{i}=\prod_{j}||\boldsymbol{\Phi}_{j}||_{F}^{2}$,
$\rho_{i}=\prod_{j}||\boldsymbol{\Psi}_{j}\mathbf{A}_{j}^{T}||_{F}^{2}$. 

With the gradient obtained, we can solve (\ref{eq:Approach III})
by alternatively updating $\boldsymbol{\Phi}_{1}$ and $\boldsymbol{\Phi}_{2}$
as follows: 
\begin{equation}
\boldsymbol{\Phi}_{i}^{(t+1)}=\boldsymbol{\Phi}_{i}^{(t)}-\eta\frac{\partial f}{\partial\boldsymbol{\Phi}_{i}},
\end{equation}
where $\eta>0$ is a step size parameter. The algorithm for solving
(\ref{eq:Approach III}) is summarized in Algorithm 2.

\begin{algorithm}[h]
\caption{\textbf{Design Approach II}}

\textbf{Input: }$\boldsymbol{\Psi}_{i}$ $(i=1,\ 2)$, $\boldsymbol{\Phi}_{i}^{(0)}$
$(i=1,\ 2)$, $\alpha$, $\beta$, $\eta$, $t=0$.

\textbf{Output:} $\hat{\boldsymbol{\Phi}}_{i}$ $(i=1,\ 2)$.

1:\textbf{ Repeat}

2: \hspace{1em}\textbf{for} $i=1,\ 2$ \textbf{do}

3:\hspace{2em} $\boldsymbol{\Phi}_{i}^{(t+1)}=\boldsymbol{\Phi}_{i}^{(t)}-\eta\frac{\partial f}{\partial\boldsymbol{\Phi}_{i}},$
where $\frac{\partial f}{\partial\boldsymbol{\Phi}_{i}}$ is given
by (\ref{eq:gradient_2D});

4:\hspace{1em} \textbf{end}

5:\hspace{1em} $t=t+1$;

6:\textbf{ Until} a stopping criteria is met.

7:\textbf{ }Normalization for $i=1,\ 2$: $\hat{\boldsymbol{\Phi}}_{i}=\sqrt{N_{i}}\boldsymbol{\Phi}_{i}/||\boldsymbol{\Phi}_{i}||_{F}$.
\end{algorithm}

Till now, we have considered optimizing the multidimensional sensing
matrix when the sparsifying dictionaries for each tensor mode are
given. For the purpose of joint optimization, we will proceed to optimize
the dictionaries by coupling fixed sensing matrices. The joint optimization
will eventually be achieved by alternatively optimizing the sensing
matrices and the sparsifying dictionaries.

\section{Jointly Learning The Multidimensional Dictionary and Sensing Matrix}

\label{sec:PsiDesign}

In this section, we first propose a sensing-matrix-coupled method
for multidimensional sparsifying dictionary learning. Then it is combined
with the previously introduced optimization approach for a multilinear
sensing matrix to yield a joint optimization algorithm. In the spirit
of the coupled KSVD method \cite{Duarte2009}, our approach for dictionary
learning can be viewed as a sensing-matrix-coupled version of a tensor
KSVD algorithm. We start by briefly introducing the coupled KSVD method.

\subsection{Coupled KSVD}

\label{sub:cKSVD and KHOSVD}

The Coupled KSVD (cKSVD) \cite{Duarte2009} is a dictionary learning
approach for vectorized signals. Let $\mathbf{X}=\left[\begin{array}{ccc}
\mathbf{x}_{1} & ... & \mathbf{x}_{T}\end{array}\right]$ be a $N\times T$ matrix containing a training sequence of $T$ signals
$\mathbf{x}_{1},...,\mathbf{x}_{T}$. The cKSVD aims to solve the
following problem, i.e., to learn a dictionary $\boldsymbol{\Psi}\in\mathbb{R}^{N\times\hat{N}}$
from $\mathbf{X}$:
\begin{equation}
\underset{\boldsymbol{\Psi},\mathbf{S}}{min}\ \gamma||\mathbf{X}-\boldsymbol{\Psi}\mathbf{S}||_{F}^{2}+||\mathbf{Y}-\boldsymbol{\Phi\Psi}\mathbf{S}||_{F}^{2},\ s.t.\ \forall i,\ ||\mathbf{s}_{i}||_{0}\leq K,\label{eq:coupled KSVD}
\end{equation}
where $\mathbf{S}=\left[\begin{array}{ccc}
\mathbf{s}_{1} & ... & \mathbf{s}_{T}\end{array}\right]$ is the sparse representation with size $\hat{N}\times T$, $\gamma>0$
is a tuning parameter and $\mathbf{Y}\in\mathbb{R}^{M\times T}$ contains
the measurement vectors taken by the sensing matrix $\boldsymbol{\Phi}\in\mathbb{R}^{M\times N}$,
i.e., $\mathbf{Y}=\left[\begin{array}{ccc}
\mathbf{y}_{1} & ... & \mathbf{y}_{T}\end{array}\right]$ and $\mathbf{Y}=\boldsymbol{\Phi}\mathbf{X}+\mathbf{E}$ with $\mathbf{E}\in\mathbb{R}^{M\times T}$
representing the noise. Then the problem in (\ref{eq:coupled KSVD})
is reformatted as:
\begin{equation}
\underset{\boldsymbol{\Psi},\mathbf{S}}{min}\ ||\mathbf{Z}-\mathbf{D}\mathbf{S}||_{F}^{2},\ s.t.\ \forall i,\ ||\mathbf{s}_{i}||_{0}\leq K,\label{eq:eqCoupled KSVD}
\end{equation}
where $\mathbf{Z}=\left[\begin{array}{cc}
\gamma\mathbf{X}^{T} & \mathbf{Y}^{T}\end{array}\right]^{T}$, $\mathbf{D}=\left[\begin{array}{cc}
\gamma\mathbf{I}_{N} & \boldsymbol{\Phi}^{T}\end{array}\right]^{T}\boldsymbol{\Psi}$. The problem can then be solved following the conventional KSVD algorithm
\cite{Aharon2006} and conducting proper normalization. 

Specifically, with an initial arbitrary $\boldsymbol{\Psi}$, it first
recovers $\mathbf{S}$ using some available algorithms, e.g., OMP.
Then the objective in (\ref{eq:eqCoupled KSVD}) is rewritten as:
\begin{equation}
\underset{\boldsymbol{\Psi},\mathbf{S}}{min}\ ||\tilde{\mathbf{R}}_{p}-\mathbf{d}_{p}\tilde{\mathbf{s}}_{p}^{T}||_{F}^{2},
\end{equation}
where $p$ is the index of the current atom we aim to update, $\tilde{\mathbf{s}}_{p}^{T}$
is the row of $\mathbf{S}$ where the zeros have been removed, $\mathbf{R}_{p}=\mathbf{Z}-\sum_{q\neq p}\mathbf{d}_{q}\mathbf{s}_{q}^{T}$
and $\tilde{\mathbf{R}}_{p}$ denotes the columns of $\mathbf{R}_{p}$
corresponding to $\tilde{\mathbf{s}}_{p}^{T}$. Let $\tilde{\mathbf{R}}_{p}=\mathbf{U}_{\mathbf{R}}\boldsymbol{\Lambda}_{\mathbf{R}}\mathbf{V}_{\mathbf{R}}^{T}$
be a SVD of $\tilde{\mathbf{R}}_{p}$, then the highest component
of the coupled error $\tilde{\mathbf{R}}_{p}$ can be eliminated by
defining: 
\begin{align}
\mathbf{\hat{\boldsymbol{\psi}}}_{p} & =(\gamma^{2}\mathbf{I}_{N}+\boldsymbol{\Phi}^{T}\boldsymbol{\Phi})^{-1}\left[\begin{array}{cc}
\gamma\mathbf{I}_{N} & \boldsymbol{\Phi}^{T}\end{array}\right]\mathbf{u}_{\mathbf{R}}^{1},\\
\tilde{\mathbf{s}}_{p} & =||\mathbf{\hat{\boldsymbol{\psi}}}_{p}||_{2}\lambda_{\mathbf{R}}^{1}\mathbf{v}_{\mathbf{R}}^{1},
\end{align}
where $\lambda_{\mathbf{R}}^{1}$ is the largest singular value of
$\tilde{\mathbf{R}}_{p}$ and $\mathbf{u}_{\mathbf{R}}^{1}$, $\mathbf{v}_{\mathbf{R}}^{1}$
are the corresponding left and right singular vectors. The update
column $p$ of $\boldsymbol{\Psi}$ is obtained after normalization:
$\mathbf{\hat{\boldsymbol{\psi}}}_{p}=\mathbf{\hat{\boldsymbol{\psi}}}_{p}/||\mathbf{\hat{\boldsymbol{\psi}}}_{p}||_{2}$.
The above process is then iterated to update every atom of $\boldsymbol{\Psi}$. 

Clearly the sensing matrix has been taken into account during the
dictionary learning process, which has been shown to be beneficial
for CS reconstruction performance \cite{Duarte2009}. In order to
learn multidimensional separable dictionaries for high dimensional
signals, and to achieve joint optimization of the multidimensional
dictionary and sensing matrix, we will derive a coupled-KSVD algorithm
for a tensor, i.e., cTKSVD, in the following section. Again for simplicity
we will still describe the main flow for 2-D signals, i.e., $n=2$.

\subsection{The cTKSVD Approach}

Consider a training sequence of 2-D signals $\mathbf{X}_{1},...,\mathbf{X}_{T}$,
we obtain a tensor $\underline{\mathbf{X}}\in\mathbb{R}^{N_{1}\times N_{2}\times T}$
by stacking them along the third dimension. Denoting the stack of
the sparse representations $\mathbf{S}_{i}\in\mathbb{R}^{\hat{N}_{1}\times\hat{N}_{2}},\ (i=1,...,T)$
by $\underline{\mathbf{S}}\in\mathbb{R}^{\hat{N}_{1}\times\hat{N}_{2}\times T}$,
we propose the following optimization problem to learn the multidimensional
dictionary:
\begin{equation}
\underset{\boldsymbol{\Psi}_{1},\boldsymbol{\Psi}_{2},\mathbf{\underline{S}}}{min}\ ||\underline{\mathbf{Z}}-\underline{\mathbf{S}}\times_{1}\mathbf{D}_{1}\times_{2}\mathbf{D}_{2}||_{F}^{2},\ s.t.,\ \forall i,\ ||\mathbf{S}_{i}||_{0}\leq K,\label{eq:Coupled K-HOSVD}
\end{equation}
in which 
\begin{align}
\underline{\mathbf{Z}} & =\left[\begin{array}{cc}
\gamma^{2}\underline{\mathbf{X}} & \gamma\underline{\mathbf{Y}}_{2}\\
\gamma\underline{\mathbf{Y}}_{1} & \underline{\mathbf{Y}}
\end{array}\right],\ \underline{\mathbf{Y}}_{i}=\underline{\mathbf{X}}\times_{i}\boldsymbol{\Phi}_{i}+\underline{\mathbf{E}}_{i},\\
\mathbf{D}_{1} & =\left[\begin{array}{c}
\gamma\mathbf{I}_{\hat{N}_{1}}\\
\boldsymbol{\Phi}_{1}
\end{array}\right]\boldsymbol{\Psi}_{1},\ \mathbf{D}_{2}=\left[\begin{array}{c}
\gamma\mathbf{I}_{\hat{N}_{2}}\\
\boldsymbol{\Phi}_{2}
\end{array}\right]\boldsymbol{\Psi}_{2},
\end{align}
and $\gamma>0$ is a tuning parameter.

The problem in (\ref{eq:Coupled K-HOSVD}) aims to minimize the representation
error $||\underline{\mathbf{X}}-\underline{\mathbf{S}}\times_{1}\boldsymbol{\Psi}_{1}\times_{2}\boldsymbol{\Psi}_{2}||_{F}^{2}$
and the overall projection error $||\underline{\mathbf{Y}}-\underline{\mathbf{S}}\times_{1}\mathbf{A}_{1}\times_{2}\mathbf{A}_{2}||_{F}^{2}$
with constraints on the sparsity of each slice of the tensor. In addition,
it also takes into account the projection errors induced by $\boldsymbol{\Phi}_{1}$
and $\boldsymbol{\Phi}_{2}$ individually. 

Using an available sparse reconstruction algorithm for the TCS, e.g.,
Tensor OMP (TOMP) \cite{Caiafa:2013}, and initial dictionaries $\boldsymbol{\Psi}_{1},\ \boldsymbol{\Psi}_{2}$,
the sparse representation $\underline{\mathbf{S}}$ can be estimated
first. Then we update the multilinear dictionary alternately. We first
update the atoms of $\boldsymbol{\Psi}_{1}$ with $\boldsymbol{\Psi}_{2}$
fixed. The objective in (\ref{eq:Coupled K-HOSVD}) is rewritten as:
\begin{equation}
||\underline{\mathbf{R}}_{p_{1}}-\sum_{q_{2}}(\mathbf{d}_{1})_{p_{1}}\circ(\mathbf{d}_{2})_{q_{2}}\circ\mathbf{s}_{(p_{1}-1)\hat{N}_{2}+q_{2}}||_{F}^{2},\label{eq:ErrorKHOSVD}
\end{equation}
where $\underline{\mathbf{R}}_{p_{1}}=\underline{\mathbf{Z}}-\sum_{q_{1}\neq p_{1}}\sum_{q_{2}}(\mathbf{d}_{1})_{q_{1}}\circ(\mathbf{d}_{2})_{q_{2}}\circ\mathbf{s}_{(q_{1}-1)\hat{N}_{2}+q_{2}}$;
$p_{1}$ is the index of the atom for the current update and $q_{1},\ q_{2}$
denote the indices of the remaining atoms of $\boldsymbol{\Psi}_{1}$
and all the atoms of $\boldsymbol{\Psi}_{2}$, respectively; $\mathbf{d}_{1}$,
$\mathbf{d}_{2}$ are columns of $\mathbf{D}_{1}$, $\mathbf{D}_{2}$;
$\mathbf{s}$ is the mode-3 vector of $\underline{\mathbf{S}}$. Then
to satisfy the sparsity constraint in (\ref{eq:Coupled K-HOSVD}),
we only keep the non-zero entries of $\mathbf{s}_{(p_{1}-1)\hat{N}_{2}+q_{2}}$
and the corresponding subset of $\underline{\mathbf{R}}_{p_{1}}$
to obtain: 
\begin{equation}
||\tilde{\underline{\mathbf{R}}}_{p_{1}}-\sum_{q_{2}}(\mathbf{d}_{1})_{p_{1}}\circ(\mathbf{d}_{2})_{q_{2}}\circ\mathbf{\tilde{s}}_{(p_{1}-1)\hat{N}_{2}+q_{2}}||_{F}^{2}.\label{eq:ErrorKHOSVD2}
\end{equation}

Assuming that after carrying out a Higher Order SVD (HOSVD) \cite{De2000}
for $\tilde{\underline{\mathbf{R}}}_{p_{1}}$, the largest singular
value is $\lambda_{\mathbf{R}}^{1}$ and the corresponding singular
vectors are $\mathbf{u}_{\mathbf{R}}^{1}$, $\mathbf{v}_{\mathbf{R}}^{1}$
and $\boldsymbol{\omega}_{\mathbf{R}}^{1}$, we eliminate the largest
error by: 
\begin{equation}
(\hat{\mathbf{d}}_{1})_{p_{1}}=\mathbf{u}_{\mathbf{R}}^{1},\ \mathbf{D}_{2}\tilde{\mathbf{S}}_{p_{1},:,:}=\mathbf{v}_{\mathbf{R}}^{1}\circ(\lambda_{\mathbf{R}}^{1}\boldsymbol{\omega}_{\mathbf{R}}^{1}),\label{eq:update_Middle}
\end{equation}
where $\tilde{\mathbf{S}}_{p_{1},:,:}$ denotes the horizontal slice
of $\underline{\mathbf{S}}$ at index $p_{1}$ that contains only
non-zero mode-2 vectors. The atom of $\boldsymbol{\Psi}_{1}$ is then
calculated using the pseudo-inverse as: 
\begin{align}
(\mathbf{\hat{\boldsymbol{\psi}}}_{1})_{p_{1}} & =(\gamma^{2}\mathbf{I}_{N_{1}}+\boldsymbol{\Phi}_{1}^{T}\boldsymbol{\Phi}_{1})^{-1}\left[\begin{array}{cc}
\gamma\mathbf{I}_{N_{1}} & \boldsymbol{\Phi}_{1}^{T}\end{array}\right]\mathbf{u}_{\mathbf{R}}^{1}.\label{eq:updatePsi}
\end{align}
The current update is then obtained after normalization: 
\begin{align}
(\mathbf{\hat{\boldsymbol{\psi}}}_{1})_{p_{1}} & =\frac{(\mathbf{\hat{\boldsymbol{\psi}}}_{1})_{p_{1}}}{||(\mathbf{\hat{\boldsymbol{\psi}}}_{1})_{p_{1}}||_{2}},\\
\mathbf{D}_{2}\tilde{\mathbf{S}}_{p_{1},:,:} & =||(\mathbf{\hat{\boldsymbol{\psi}}}_{1})_{p_{1}}||_{2}\mathbf{v}_{\mathbf{R}}^{1}\circ(\lambda_{\mathbf{R}}^{1}\boldsymbol{\omega}_{\mathbf{R}}^{1}).\label{eq:UpdateS}
\end{align}
 Since $\mathbf{D}_{2}$ and the support indices of each mode-2 vector
in $\tilde{\mathbf{S}}_{p_{1},:,:}$ are known, the updated coefficients
$\tilde{\mathbf{S}}_{p_{1},:,:}$ can be easily calculated by the
Least Square (LS) solution. The above process is repeated for all
the atoms to update the dictionary $\boldsymbol{\Psi}_{1}$.

The next step is to update $\boldsymbol{\Psi}_{2}$ with the obtained
$\boldsymbol{\Psi}_{1}$ fixed. It follows a similar procedure to
that described previously. Specifically, the objective in (\ref{eq:Coupled K-HOSVD})
is rewritten as: 
\begin{equation}
||\tilde{\underline{\mathbf{R}}}_{p_{2}}-\sum_{q_{1}}(\mathbf{d}_{1})_{q_{1}}\circ(\mathbf{d}_{2})_{p_{2}}\circ\mathbf{\tilde{s}}_{(q_{1}-1)\hat{N}_{2}+p_{2}}||_{F}^{2},\label{eq:ErrorKHOSVD_Psi2}
\end{equation}
where $\tilde{\mathbf{s}}$ is the mode-3 vector with only non-zero
entries, $\tilde{\underline{\mathbf{R}}}_{p_{2}}$ is the corresponding
subset of $\underline{\mathbf{R}}_{p_{2}}$, $\underline{\mathbf{R}}_{p_{2}}=\underline{\mathbf{Z}}-\sum_{q_{1}}\sum_{q_{2}\neq p_{2}}(\mathbf{d}_{1})_{q_{1}}\circ(\mathbf{d}_{2})_{q_{2}}\circ\mathbf{s}_{(q_{1}-1)\hat{N}_{2}+q_{2}}$
and $p_{2}$ is the index of the atom for current update. A HOSVD
is carried out for $\tilde{\underline{\mathbf{R}}}_{p_{2}}$ and the
update steps corresponding to (\ref{eq:update_Middle}) - (\ref{eq:UpdateS})
now become: 
\begin{align}
(\hat{\mathbf{d}}_{2})_{p_{2}} & =\mathbf{v}_{\mathbf{R}}^{1},\ \mathbf{D}_{1}\tilde{\mathbf{S}}_{:,p_{2},:}=\mathbf{u}_{\mathbf{R}}^{1}\circ(\lambda_{\mathbf{R}}^{1}\boldsymbol{\omega}_{\mathbf{R}}^{1}),\\
(\mathbf{\hat{\boldsymbol{\psi}}}_{2})_{p_{2}} & =(\gamma^{2}\mathbf{I}_{N_{2}}+\boldsymbol{\Phi}_{2}^{T}\boldsymbol{\Phi}_{2})^{-1}\left[\begin{array}{cc}
\gamma\mathbf{I}_{N_{2}} & \boldsymbol{\Phi}_{2}^{T}\end{array}\right]\mathbf{v}_{\mathbf{R}}^{1},\label{eq:UpdatePsi2}\\
(\mathbf{\hat{\boldsymbol{\psi}}}_{2})_{p_{2}} & =(\mathbf{\hat{\boldsymbol{\psi}}}_{2})_{p_{2}}/||(\mathbf{\hat{\boldsymbol{\psi}}}_{2})_{p_{2}}||_{2},\\
\mathbf{D}_{1}\tilde{\mathbf{S}}_{:,p_{2},:} & =||(\mathbf{\hat{\boldsymbol{\psi}}}_{2})_{p_{2}}||_{2}\mathbf{u}_{\mathbf{R}}^{1}\circ(\lambda_{\mathbf{R}}^{1}\boldsymbol{\omega}_{\mathbf{R}}^{1}),\label{eq:UpdateS2}
\end{align}
in which $\tilde{\mathbf{S}}_{:,p_{2},:}$ represents the lateral
slice at index $p_{2}$ and its updated elements can also be calculated
using LS. The dictionary $\boldsymbol{\Psi}_{2}$ is then updated
iteratively. The whole process of updating $\underline{\mathbf{S}},\ \boldsymbol{\Psi}_{1},\ \boldsymbol{\Psi}_{2}$
is repeated to obtain the final solution of (\ref{eq:Coupled K-HOSVD}).

The uncoupled version of the proposed cTKSVD method (denoted by TKSVD)
can be easily obtained by modifying the problem in (\ref{eq:Coupled K-HOSVD})
to: 
\begin{equation}
\underset{\boldsymbol{\Psi}_{1},\boldsymbol{\Psi}_{2},\mathbf{\underline{S}}}{min}\ ||\underline{\mathbf{X}}-\underline{\mathbf{S}}\times_{1}\boldsymbol{\Psi}_{1}\times_{2}\boldsymbol{\Psi}_{2}||_{F}^{2},\ s.t.\ \forall i,\ ||\mathbf{S}_{i}||_{0}\leq K,\label{eq:KHOSVD obj}
\end{equation}
and it can be solved following the same procedures as described previously
for cTKSVD except that the steps of pseudo-inverse and normalization
are no longer needed. 

The proposed cTKSVD for multidimensional dictionary learning is different
to the KHOSVD method \cite{Roemer2014}, i.e., another tensor-based
dictionary learning approach obtained by extending the KSVD method.
The learning process of KHOSVD follows the same train of thought as
with the conventional KSVD method, except that to eliminate the largest
error in each iteration, a HOSVD \cite{De2000}, i.e., SVD for tensors,
is employed. However, the process of KHOSVD does not take full advantage
of the multilinear structure and involves duplicated updating of the
atoms, which leads to a slow convergence speed. The proposed cTKSVD
approach is distinct from KHOSVD in the following respects. First,
during the update of each atom, a slice of the coefficient is updated
accordingly in cTKSVD; while only a vector is updated in KHOSVD. Therefore,
in cTKSVD, each iteration of the outer loop contains $\hat{N}_{1}+\hat{N}_{2}$
inner iterations, which is $\hat{N}_{1}\hat{N}_{2}$ for KHOSVD (and
for KSVD). It means that cTKSVD requires HOSVD to be executed $\hat{N}_{1}\hat{N}_{2}-\hat{N}_{1}-\hat{N}_{2}$
fewer times than for the KHOSVD method and hence reduces the complexity.
In addition, KHOSVD does not take into account the influence from
the sensing matrix. The benefit of coupling of the sensing matrices
in cTKSVD will be shown by simulations in Section \ref{sub:Simulation_Psi}.

Here, we also provide the problem formulation when one needs to learn
3-D sparsifying dictionaries. The cTKSVD for cases where $n>2$ can
be modeled following a similar strategy. For a training sequence consisting
of $T$ stacked 3-D signals $\underline{\mathbf{X}}\in\mathbb{R}^{N_{1}\times N_{2}\times N_{3}\times T}$,
we learn the dictionaries by solving: 
\begin{equation}
\underset{\boldsymbol{\Psi}_{1},\boldsymbol{\Psi}_{2},\boldsymbol{\Psi}_{3},\mathbf{\underline{S}}}{min}\ ||\underline{\mathbf{Z}}-\underline{\mathbf{S}}\times_{1}\mathbf{D}_{1}\times_{2}\mathbf{D}_{2}\times_{3}\mathbf{D}_{3}||_{F}^{2},\ s.t.,\ \forall i,\ ||\mathbf{\underline{S}}_{i}||_{0}\leq K,
\end{equation}
in which 
\begin{align}
\underline{\mathbf{Z}} & =\left[\begin{array}{cc}
\gamma^{2}\underline{\mathbf{G}}_{1} & \gamma\underline{\mathbf{G}}_{2}\\
\gamma\underline{\mathbf{G}}_{3} & \underline{\mathbf{G}}_{4}
\end{array}\right],\ \mathbf{D}_{1}=\left[\begin{array}{c}
\gamma\mathbf{I}_{\hat{N}_{1}}\\
\boldsymbol{\Phi}_{1}
\end{array}\right]\boldsymbol{\Psi}_{1},\\
\mathbf{D}_{2} & =\left[\begin{array}{c}
\gamma\mathbf{I}_{\hat{N}_{2}}\\
\boldsymbol{\Phi}_{2}
\end{array}\right]\boldsymbol{\Psi}_{2},\ \mathbf{D}_{3}=\left[\begin{array}{c}
\gamma\mathbf{I}_{\hat{N}_{3}}\\
\boldsymbol{\Phi}_{3}
\end{array}\right]\boldsymbol{\Psi}_{3},
\end{align}
and if we denote the operator ``$\nearrow_{3}$'' as stacking tensors
along their third mode, then in the above formulation of $\underline{\mathbf{Z}}$,
\begin{align}
\underline{\mathbf{G}}_{1} & =(\gamma\underline{\mathbf{X}})\nearrow_{3}(\underline{\mathbf{Y}}_{3}),\ \underline{\mathbf{G}}_{2}=(\gamma\underline{\mathbf{Y}}_{2})\nearrow_{3}(\underline{\mathbf{Y}}_{23}),\nonumber \\
\underline{\mathbf{G}}_{3} & =(\gamma\underline{\mathbf{Y}}_{1})\nearrow_{3}(\underline{\mathbf{Y}}_{13}),\ \underline{\mathbf{G}}_{4}=(\gamma\underline{\mathbf{Y}})\nearrow_{3}(\underline{\mathbf{Y}}_{12}),\nonumber \\
\underline{\mathbf{Y}}_{i} & =\underline{\mathbf{X}}\times_{i}\boldsymbol{\Phi}_{i}+\underline{\mathbf{E}}_{i},\ \underline{\mathbf{Y}}_{ij}=\underline{\mathbf{X}}\times_{i}\boldsymbol{\Phi}_{i}\times_{j}\boldsymbol{\Phi}_{j}+\underline{\mathbf{E}}_{ij}.
\end{align}
The problem can then be solved following similar steps to those introduced
earlier in this section.

We have now derived the method of learning the sparsifying dictionaries
when the multilinear sensing matrix is fixed. Combining this approach
with the methods of optimizing the sensing matrices elaborated in
Section \ref{sub:SensingDesign}, we can then jointly optimize $\boldsymbol{\Phi}_{1},\ \boldsymbol{\Phi}_{2}$
and $\boldsymbol{\Psi}_{1},\ \boldsymbol{\Psi}_{2}$ by alternating
between them. The overall procedure is summarized in Algorithm 3.

\begin{algorithm}[h]
\caption{\textbf{Joint Optimization} }
\textbf{Input: }$\boldsymbol{\Psi}_{i}^{(0)}$ $(i=1,\ 2)$, $\boldsymbol{\Phi}_{i}^{(0)}$
$(i=1,\ 2)$, $\underline{\mathbf{X}}$, $\alpha$, $\beta$, $\eta$,
$\gamma$, 

\hspace{3.1em}$iter=0$.

\textbf{Output:} $\hat{\boldsymbol{\Phi}}_{i}$ $(i=1,\ 2)$, $\hat{\boldsymbol{\Psi}}_{i}$
$(i=1,\ 2)$.

1:\textbf{ Repeat until convergence:}

2:\hspace{1.3em}For $\hat{\boldsymbol{\Psi}}_{i}^{(iter)}$ $(i=1,\ 2)$
fixed, optimize $\hat{\boldsymbol{\Phi}}_{i}^{(iter+1)}$ $(i=$

\hspace{2.1em}$1,\ 2)$ using one of the approaches given in Section
\ref{sub:SensingDesign};

3:\hspace{1.3em}For $\hat{\boldsymbol{\Psi}}_{i}^{(iter)}$ , $\hat{\boldsymbol{\Phi}}_{i}^{(iter+1)}$
$(i=1,\ 2)$ fixed, solve (\ref{eq:Coupled K-HOSVD}) using 

\hspace{2.1em}TOMP to obtain $\underline{\hat{\mathbf{S}}}$;

4:\hspace{1.3em}\textbf{For $p_{1}=1$ }to\textbf{ $\hat{N}_{1}$}

5:\hspace{2.3em}Compute $\tilde{\underline{\mathbf{R}}}_{p_{1}}$
using (\ref{eq:Coupled K-HOSVD}) - (\ref{eq:ErrorKHOSVD});

6:\hspace{2.3em}Do HOSVD to $\tilde{\underline{\mathbf{R}}}_{p_{1}}$
to obtain $\lambda_{\mathbf{R}}^{1}$, $\mathbf{u}_{\mathbf{R}}^{1}$,
$\mathbf{v}_{\mathbf{R}}^{1}$ and $\boldsymbol{\omega}_{\mathbf{R}}^{1}$;

7:\hspace{2.3em}Update $(\mathbf{\hat{\boldsymbol{\psi}}}_{1}^{(iter+1)})_{p_{1}}$,
$\mathbf{D}_{2}\mathbf{\tilde{S}}_{p_{1,:,:}}$ using (\ref{eq:updatePsi})
- (\ref{eq:UpdateS}) and 

\hspace{3.1em}calculate $\mathbf{\tilde{S}}_{p_{1,:,:}}$ by LS;

8:\hspace{1.3em}\textbf{end}

9:\hspace{1.3em}\textbf{For $p_{2}=1$ }to\textbf{ $\hat{N}_{2}$}

10:\hspace{1.8em}Compute $\tilde{\underline{\mathbf{R}}}_{p_{2}}$
using (\ref{eq:Coupled K-HOSVD}) and (\ref{eq:ErrorKHOSVD_Psi2});

11:\hspace{1.8em}Do HOSVD to $\tilde{\underline{\mathbf{R}}}_{p_{2}}$
to obtain $\lambda_{\mathbf{R}}^{1}$, $\mathbf{u}_{\mathbf{R}}^{1}$,
$\mathbf{v}_{\mathbf{R}}^{1}$ and $\boldsymbol{\omega}_{\mathbf{R}}^{1}$;

12:\hspace{1.8em}Update $(\mathbf{\hat{\boldsymbol{\psi}}}_{2}^{(iter+1)})_{p_{2}}$,
$\mathbf{D}_{1}\mathbf{\tilde{S}}_{:,p_{2},:}$ using (\ref{eq:UpdatePsi2})
- (\ref{eq:UpdateS2}) and 

\hspace{3.1em}calculate $\mathbf{\tilde{S}}_{:,p_{2},:}$ by LS;

13:\hspace{0.8em}\textbf{end}

14:\hspace{0.8em}$iter=iter+1$; 
\end{algorithm}

\section{Experimental Results}

\label{sec:simulations}

In this section, we evaluate the proposed approaches via simulations
using both synthetic data and real images. We first test the sensing
matrix design approaches proposed in Section \ref{sub:SensingDesign}
with the sparsifying dictionaries being given. Then the cTKSVD approach
is evaluated when the sensing matrices are fixed. Finally the experiments
for the joint optimization of the two are presented.

\subsection{Optimal Multidimensional Sensing Matrix}

\label{sub:Simulation_Phi}

This section is intended to examine the proposed separable approach
I and non-separable approach II for multidimensional sensing matrix
design. Before doing so, we first test the tuning parameters for Approach
II, i.e., the non-separable design approach presented in Section \ref{sub:SensingDesign}-2.
As detailed in Section \ref{sub:SensingDesign}-1, Approach I has
a closed form solution and there are no tuning parameters involved. 

We evaluate the Mean Squared Error (MSE) performance of different
sensing matrices generated using Approach II with various parameters
and the results are reported by averaging over 500 trials. A random
2D signal $\mathbf{S}\in\mathbb{R}^{64\times64}$ with sparsity $K=80$
is generated, where the randomly placed non-zero elements follow an
i.i.d zero-mean unit-variance Gaussian distribution. Both the dictionaries
$\boldsymbol{\Psi}_{i}\in\mathbb{R}^{64\times256}\ (i=1,\ 2)$ and
the initial sensing matrices $\boldsymbol{\Phi}_{i}\in\mathbb{R}^{40\times64}\ (i=1,\ 2)$
are generated randomly with i.i.d zero-mean unit-variance Gaussian
distributions, and the dictionaries are then column normalized while
the sensing matrices are normalized by: $\boldsymbol{\Phi}_{i}=\sqrt{64}\boldsymbol{\Phi}_{i}/||\boldsymbol{\Phi}_{i}||_{F}.$
When taking measurements, random additive Gaussian noise with variance
$\sigma^{2}$ is induced. A constant step size $\eta=1e-7$ is used
for Approach II and the BP solver SPGL1 \cite{Berg2008} is employed
for reconstructions. 
\begin{figure}
\centering{}\includegraphics{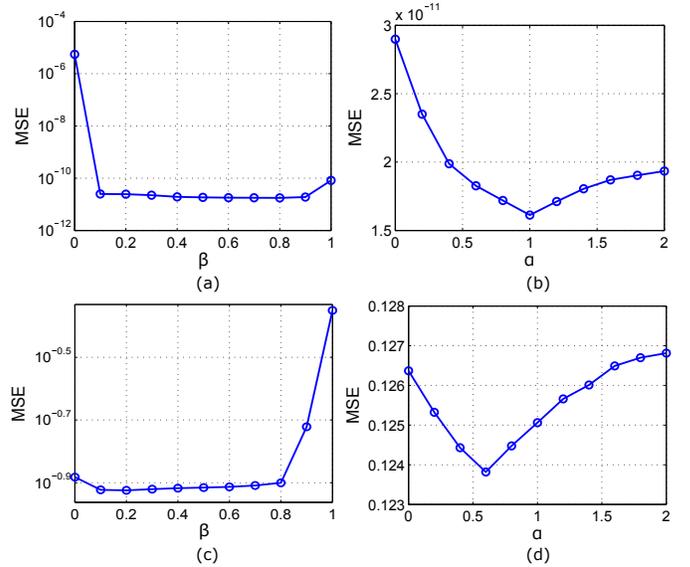}\caption{MSE performance of sensing matrices generated by Approach II with
different values of $\alpha$ and $\beta$. (a) $\sigma^{2}=0,\ \alpha=1;$
(b) $\sigma^{2}=0,\ \beta=0.8;$ (c) $\sigma^{2}=10^{-2},\ \alpha=1;$
(d) $\sigma^{2}=10^{-2},\ \beta=0.2.$\label{fig:Phi_parameters}}
\end{figure}

Fig. \ref{fig:Phi_parameters} illustrates the results for the parameter
tests. In Fig. \ref{fig:Phi_parameters} (a) and (c), the parameter
$\beta$ is evaluated for the noiseless ($\sigma^{2}=0$) and high
noise ($\sigma^{2}=10^{-2}$) cases, respectively, when $\alpha=1$.
From both (a) and (c), we can see that when $\beta=0$ or 1, the MSE
is larger than that for the other values, which means that both terms
of Approach II that are controlled by $\beta$ are essential for obtaining
optimal sensing matrices. In addition, we can see that when $\beta$
becomes larger in the range of $[0.1,\ 0.9]$, the MSE decreases slightly
in (a), but increases slightly in (b). This indicates the choice of
$\beta$ under different conditions of sensing noise, which is consistent
with that observed in \cite{Bai2015}. Thus in the remaining experiments,
we take $\beta=0.8$ when sensing noise is low and $\beta=0.2$ when
the noise is high. Fig. \ref{fig:Phi_parameters} (b) and (d) demonstrate
the MSE results for the tests of parameter $\alpha$. It is observed
that $\alpha=1$ is optimal for the noiseless case while it becomes
$\alpha=0.6$ when high noise exists. Therefore a larger $\alpha$
is preferred when low noise is involved, which needs to be reduced
accordingly when the noise becomes higher.

\begin{figure}
\begin{centering}
\subfloat[]{\begin{centering}
\includegraphics{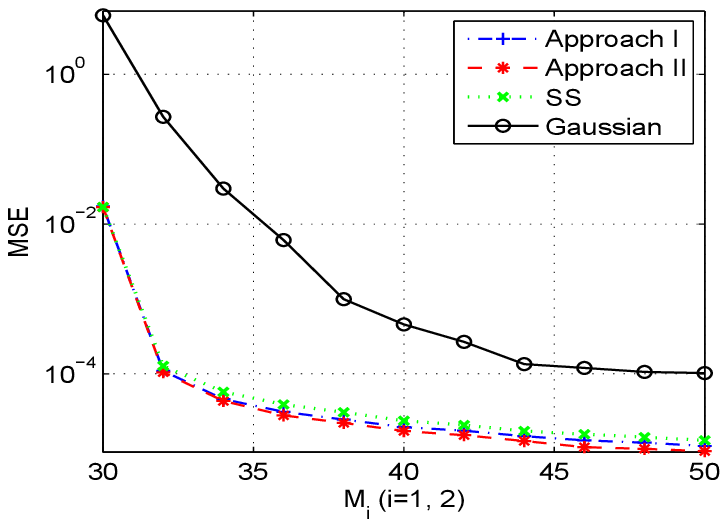}
\par\end{centering}

}
\par\end{centering}

\centering{}\subfloat[]{\begin{centering}
\includegraphics{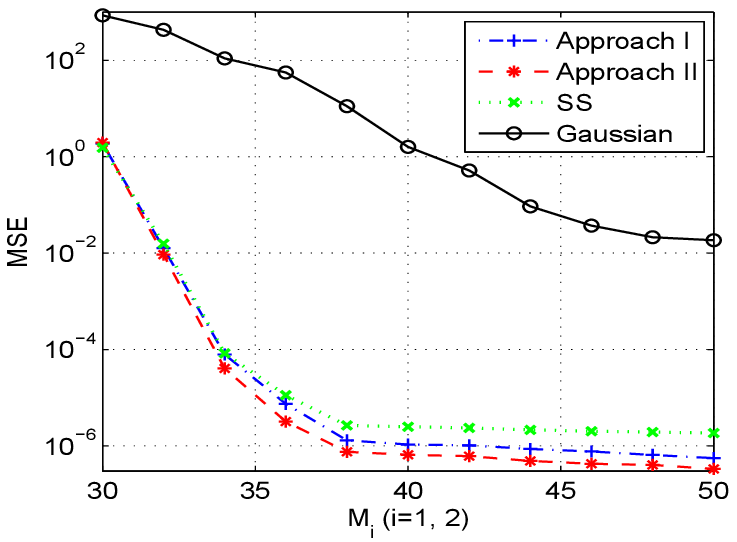}
\par\end{centering}

}\caption{MSE performance of different sensing matrices for (a) the BP, (b)
the OMP when $M_{i}\ (i=1,\ 2)$ varies. ($K=80,\ N_{1}=N_{2}=64,\ \hat{N}_{1}=\hat{N}_{2}=256$
and $\sigma^{2}=10^{-4}$) \label{fig:Phi_M}}
\end{figure}
\begin{figure}
\begin{centering}
\subfloat[]{\begin{centering}
\includegraphics{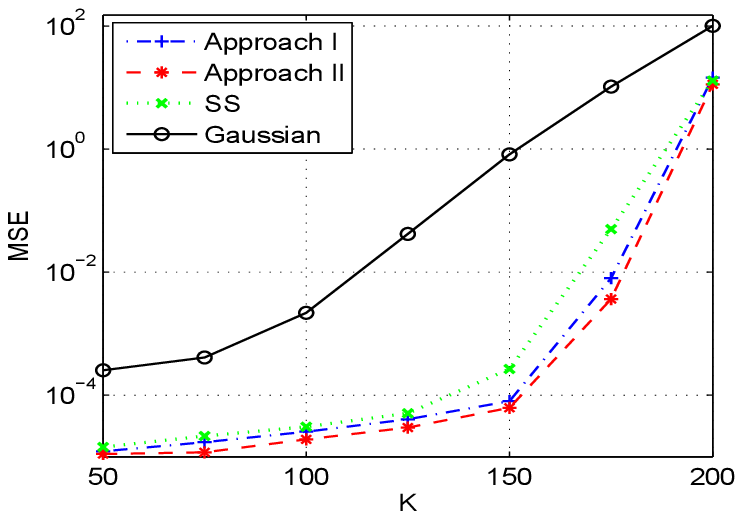}
\par\end{centering}

}
\par\end{centering}

\centering{}\subfloat[]{\begin{centering}
\includegraphics{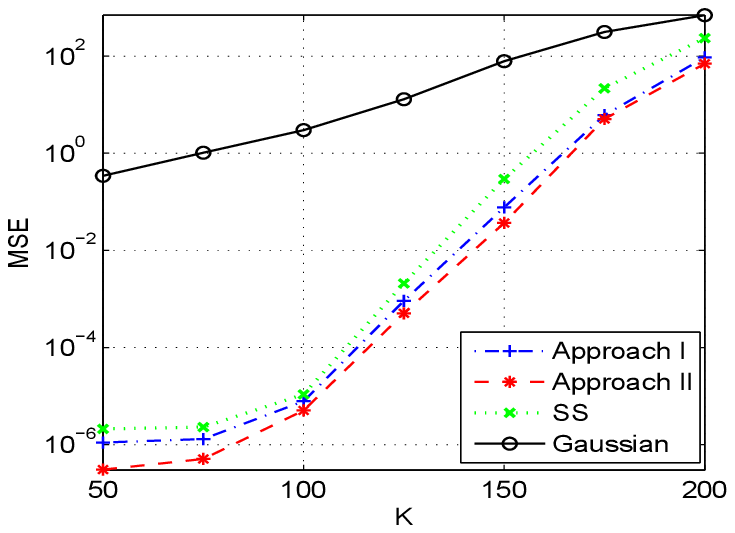}
\par\end{centering}

}\caption{MSE performance of different sensing matrices for (a) the BP, (b)
the OMP when $K$ varies. ($M_{1}=M_{2}=40,\ N_{1}=N_{2}=64,\ \hat{N}_{1}=\hat{N}_{2}=256$
and $\sigma^{2}=10^{-4}$) \label{fig:Phi_K}}
\end{figure}

We then proceed to examine the performance of both the proposed approaches.
As this is the first work to optimize the multidimensional sensing
matrix, we take the i.i.d Gaussian sensing matrices that are commonly
used in CS problems for comparison. Besides, since Sapiro's approach
\cite{Duarte2009} has the same spirit to that of Approach I (as reviewed
in Section \ref{sub:CS sensing design}), it can be easily extended
to the multidimensional case, i.e., individually generating $\boldsymbol{\Phi}_{i}\ (i=1,\ 2)$
using the approach in \cite{Duarte2009}. We hence also include it
in the comparisons and denote it by Separable Sapiro's approach (SS).
The previously described synthetic data is generated for the experiments
and both BP and OMP are investigated for the reconstruction.

Different sensing matrices are first evaluated using BP and OMP when
the number of measurements varies. A small amount of noise ($\sigma^{2}=10^{-4}$)
is added when taking measurements and the parameters are chosen as:
$\alpha=1,\ \beta=0.8$. From Fig. \ref{fig:Phi_M}, it can be observed
that both the proposed approaches perform much better than the Gaussian
sensing matrices, among which Approach II has better performance.
In general, the SS method performs worse than Approach I, although
the difference is not obvious at some points. Note that SS is an iterative
method while Approach I is non-iterative. 

The proposed approaches are again observed to be superior to the other
methods when the number of measurements is fixed but the signal sparsity
$K$ is varied, as shown in Fig. \ref{fig:Phi_K}. Compared to Approach
I, Approach II exhibits better performance, but at the cost of higher
computational complexity and the proper choice of the parameters.

\subsection{Optimal Multidimensional Dictionary with the Sensing Matrices Coupled}

\label{sub:Simulation_Psi}

\begin{figure}
\begin{centering}
\subfloat[]{\begin{centering}
\includegraphics{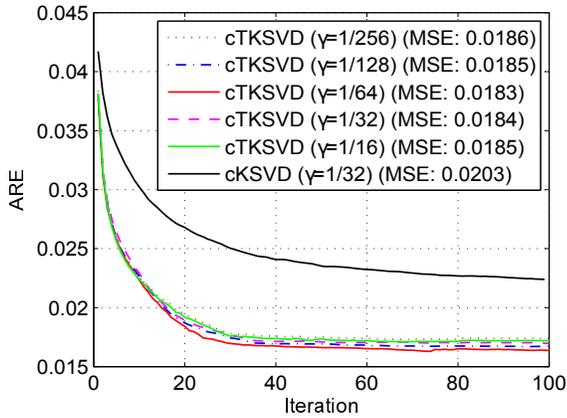}
\par\end{centering}

}
\par\end{centering}

\centering{}\subfloat[]{\begin{centering}
\includegraphics{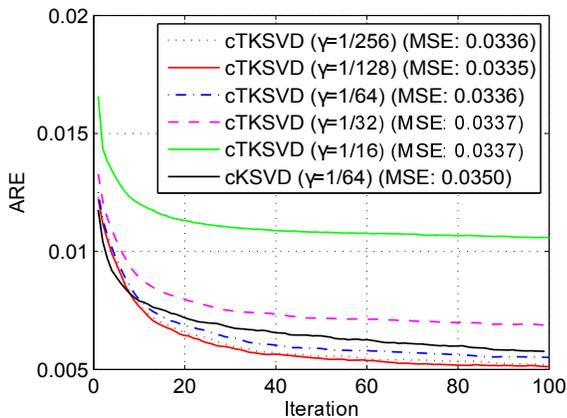}
\par\end{centering}

}\caption{Convergence behavior of cTKSVD with different values of $\gamma$
compared to that of cKSVD with its optimal parameter setting when
(a) $M_{1}=M_{2}=7;$ (b) $M_{1}=M_{2}=3$. \label{fig:Psi_parameters}}
\end{figure}

In this section, we evaluate the proposed cTKSVD method with a given
multidimensional sensing matrix. A training sequence of 5000 2D signals
($T=5000$) is generated, i.e., $\underline{\mathbf{S}}\in\mathbb{R}^{18\times18\times5000}$,
where each signal has $K=4\ (2\times2)$ randomly placed non-zero
elements that follow an i.i.d zero-mean unit-variance Gaussian distribution.
The dictionaries $\boldsymbol{\Psi}_{i}\in\mathbb{R}^{10\times18}\ (i=1,\ 2)$
are also drawn from i.i.d Gaussian distributions, followed by normalization
such that they have unit-norm columns. The time-domain training signals
$\underline{\mathbf{X}}\in\mathbb{R}^{10\times10\times5000}$ are
then formed by: $\underline{\mathbf{X}}=\underline{\mathbf{S}}\times_{1}\boldsymbol{\Psi}_{1}\times_{2}\boldsymbol{\Psi}_{2}$.
The test data of size $10\times10\times5000$ is generated following
the same procedure. Random Gaussian noise with variance $\sigma^{2}$
is added to both the training and test data. Two i.i.d random Gaussian
matrices are employed as the sensing matrices $\boldsymbol{\Phi}_{i}\in\mathbb{R}^{M_{i}\times10}\ (i=1,\ 2)$,
normalized by: $\boldsymbol{\Phi}_{i}=\sqrt{10}\boldsymbol{\Phi}_{i}/||\boldsymbol{\Phi}_{i}||_{F}.$
TOMP \cite{Caiafa2013} is utilized in both the training stage and
the reconstructions of the test stage for tensor-based approaches
and OMP is employed for the vector-based approaches.

\begin{figure}
\begin{centering}
\subfloat[]{\begin{centering}
\includegraphics{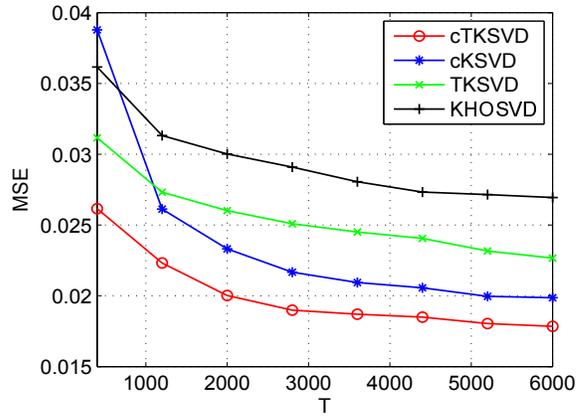}
\par\end{centering}

}
\par\end{centering}

\centering{}\subfloat[]{\begin{centering}
\includegraphics{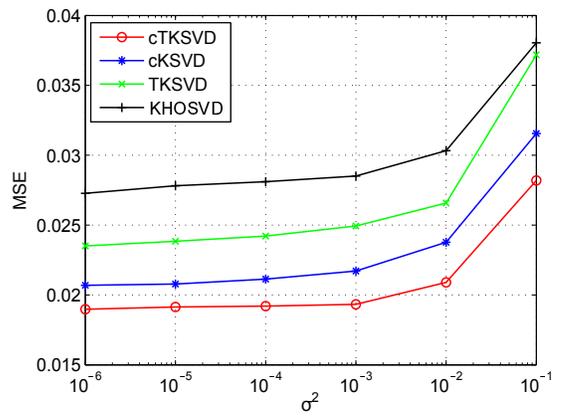}
\par\end{centering}

}\caption{MSE performance of different dictionaries when (a) $T$ varies ($\sigma^{2}=0$),
(b) $\sigma^{2}$ varies ($T=5000$). ($K=4,\ M_{1}=M_{2}=7,\ N_{1}=N_{2}=10,\ \hat{N}_{1}=\hat{N}_{2}=18$)
\label{fig:Psi_T_sigma}}
\end{figure}

We first investigate the convergence behavior of the cTKSVD approach
and examine the choice of the parameter $\gamma$. We define the Average
Representation Error (ARE) \cite{Aharon2006,Chen2013dictionary} of
cTKSVD as: $\sqrt{||\underline{\mathbf{Z}}-\underline{\mathbf{S}}\times_{1}\mathbf{D}_{1}\times_{2}\mathbf{D}_{2}||_{F}^{2}/len(\underline{\mathbf{Z}})},$
where $\underline{\mathbf{Z}}$, $\mathbf{D}_{1}$ and $\mathbf{D}_{2}$
have the same definitions as in (\ref{eq:Coupled K-HOSVD}). Fig.
\ref{fig:Psi_parameters} shows the AREs of cTKSVD at different numbers
of iterations for different values of $\gamma$. The cKSVD method
\cite{Duarte2009} (reviewed in Section \ref{sub:CS sensing design})
is also tested and only the results of the optimal $\gamma$ are displayed
in Fig. \ref{fig:Psi_parameters}. Note that cKSVD learns a single
dictionary $\boldsymbol{\Psi}\in\mathbb{R}^{100\times324}$, rather
than the separable multilinear dictionaries $\boldsymbol{\Psi}_{i}\in\mathbb{R}^{10\times18}\ (i=1,\ 2)$.
The ARE of cKSVD is thus modified accordingly as: $\sqrt{||\mathbf{Z}-\mathbf{D}\mathbf{S}||_{F}^{2}/len(\mathbf{Z})}$,
in which the symbols follow the definitions in (\ref{eq:eqCoupled KSVD}).
From Fig. \ref{fig:Psi_parameters}, it can be seen that cTKSVD exhibits
stable convergence behavior with different parameters. It converges
to a lowest ARE with $\gamma=1/64$ when $M_{i}=7$ and the optimal
$\gamma$ is $1/128$ when $M_{i}=3$. The reconstruction MSE values
are also shown in the legend, which are similar to each other but
reveal the same optimal choice of $\gamma$ as described. Thus the
optimal $\gamma$ is lower when the number of measurements decreases,
which is consistent with the observation in \cite{Duarte2009}. In
both experiments, cTKSVD with the optimal $\gamma$ outperforms cKSVD
in terms of ARE and MSE.

Then the MSE performance of dictionaries learned by cTKSVD is compared
with that of cKSVD \cite{Duarte2009} and KHOSVD \cite{Roemer2014}
when the number of training sequences $T$ and the noise variance
$\sigma^{2}$ vary. We use $\gamma=1/64$ for cTKSVD and $\gamma=1/32$
for cKSVD. To see the benefit of coupling sensing matrices, we also
evaluate the uncoupled version of the proposed approach, i.e., TKSVD,
in the experiments. The results can be found in Fig. \ref{fig:Psi_T_sigma}.
It is observable that cTKSVD outperforms all the other methods in
terms of the reconstruction MSE. The sensing-matrix-coupled approaches
(cKSVD and cTKSVD) are superior to the uncoupled approaches (TKSVD
and KHOSVD). The TKSVD method leads to smaller MSE compared to KHOSVD,
as it fully exploits the multidimensional structure. In addition,
since cKSVD is not an approach that explicitly considers a multidimensional
dictionary, it requires longer training sequences to learn the multilinear
structure from the vectorized data. As seen in Fig. \ref{fig:Psi_T_sigma}
(a), to achieve a MSE of 0.02, cTKSVD only needs 2000 training data;
while approximately 6000 is required for the cKSVD approach. For the
same reason, the performance of cKSVD degrades dramatically when the
training data is less than 1000.

\subsection{TCS with Jointly Optimized Sensing Matrix and Dictionary}

\begin{figure}
\centering{}\includegraphics{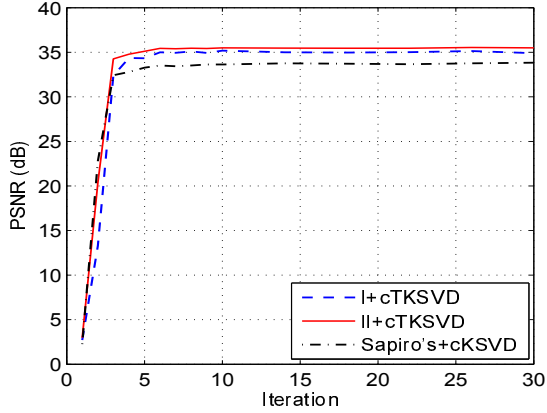}\caption{Convergence behavior of various joint optimization methods. ($T=5000,\ K=4,\ M_{1}=M_{2}=6,\ N_{1}=N_{2}=8,\ \hat{N}_{1}=\hat{N}_{2}=16,\ \sigma^{2}=0$)
\label{fig:Comb_converg}}
\end{figure}
\begin{figure}
\begin{centering}
\subfloat[]{\begin{centering}
\includegraphics{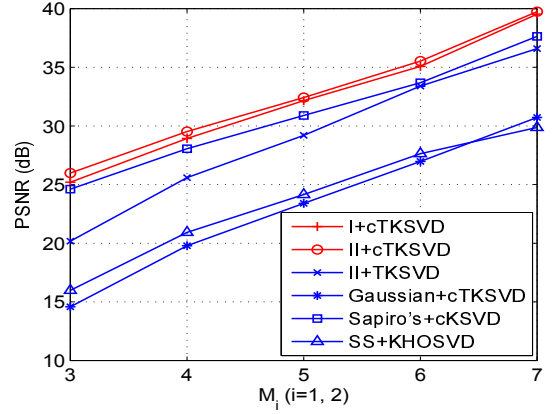}
\par\end{centering}

}
\par\end{centering}

\centering{}\subfloat[]{\begin{centering}
\includegraphics{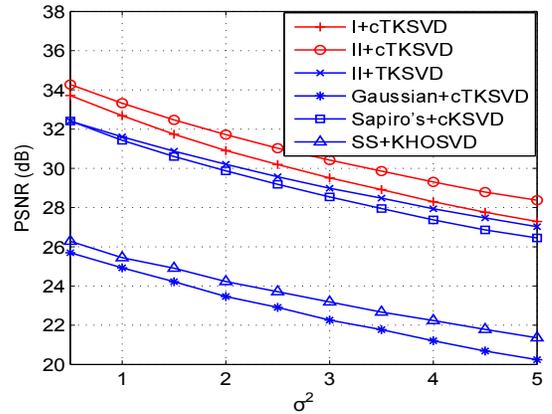}
\par\end{centering}

}\caption{PSNR performance of different methods when (a) $M_{i}\ (i=1,\ 2)$
varies ($\sigma^{2}=0$), (b) $\sigma^{2}$ varies ($M_{1}=M_{2}=6$).
($T=5000,\ K=4,\ M_{1}=M_{2}=6,\ N_{1}=N_{2}=8,\ \hat{N}_{1}=\hat{N}_{2}=16$)
\label{fig:comb_M_sigma}}
\end{figure}

Now we examine the performance of the proposed joint optimization
approach in Algorithm 3. The training data consists of 5000 $8\times8$
patches obtained by randomly extracting 25 patches from each of the
200 images in a training set from the Berkeley segmentation dataset
\cite{Martin2001}. The test data is obtained by extracting non-overlapping
$8\times8$ patches from the other 100 images in the dataset. A 2D
Discrete Cosine Transform (DCT) is employed to initialize the dictionaries
$\boldsymbol{\Psi}_{i}\in\mathbb{R}^{8\times16}\ (i=1,\ 2)$ and i.i.d
Gaussian matrices are used as the initial sensing matrices $\boldsymbol{\Phi}_{i}\in\mathbb{R}^{M_{i}\times8}\ (i=1,\ 2)$.
Random Gaussian noise with variance $\sigma^{2}$ is added to the
measurements at the test stage. We employ TOMP for reconstruction
and the Peak Signal to Noise Ratio (PSNR) is used as the evaluation
criteria.

In the first experiment, we examine the convergence behavior of Algorithm
3 when the proposed approach I and II are utilized for the sensing
matrix optimization step (respectively denote by I + cTKSVD and II
+ cTKSVD). We take $M_{1}=M_{2}=6$ and no noise is added to the measurements
at the test stage, i.e., $\sigma^{2}=0$. By conducting the simulations
performed previously to obtain the results in Fig. \ref{fig:Phi_parameters}
and \ref{fig:Psi_parameters}, the parameters are chosen as: $\alpha=3,\ \beta=0.8,\ \gamma=1/8$.
The step size for II + cTKSVD is set as: $\eta=1e-5$. The PSNR performance
for different numbers of iterations is illustrated in Fig. \ref{fig:Comb_converg}.
Since Sapiro's approach in \cite{Duarte2009} also jointly optimizes
the sensing matrix and dictionary, we include it in this figure (denoted
by Sapiro's + cKSVD). The parameter $\gamma$ is optimal at $1/2$
for cKSVD under our settings. However, note that Sapiro's approach
is only for vectorized signals in the conventional CS problem, i.e.,
a single sensing matrix $\boldsymbol{\Phi}\in\mathbb{R}^{36\times64}$
and a dictionary $\boldsymbol{\Psi}\in\mathbb{R}^{64\times256}$ are
obtained. It is not suitable for a practical TCS system, where separable
multidimensional sensing matrices $\boldsymbol{\Phi}_{i}\in\mathbb{R}^{6\times8}\ (i=1,\ 2)$
are required. Even so, from Fig. \ref{fig:Comb_converg}, we can see
the proposed approaches outperform Sapiro's approach. All the methods
converge in less than 10 iterations, among which II + cTKSVD leads
to the highest PSNR value.

Then the proposed approaches are compared with various other approaches
when the number of measurements ($M_{i}\ (i=1,\ 2)$) and the noise
variance ($\sigma^{2}$) vary. Specifically, using the notation employed
previously and by denoting the method of combining sensing matrix
design with that of the dictionary learning using a ``+'', the methods
for comparison are: II + TKSVD, Gaussian + cTKSVD, Sapiro's + cKSVD
and SS + KHOSVD. In these approaches, II + TKSVD and SS + KHOSVD are
uncoupled methods; Gaussian + cTKSVD does not involve sensing matrix
optimization; Sapiro's + cKSVD is for conventional CS system only. 

\begin{figure}
\centering{}\includegraphics{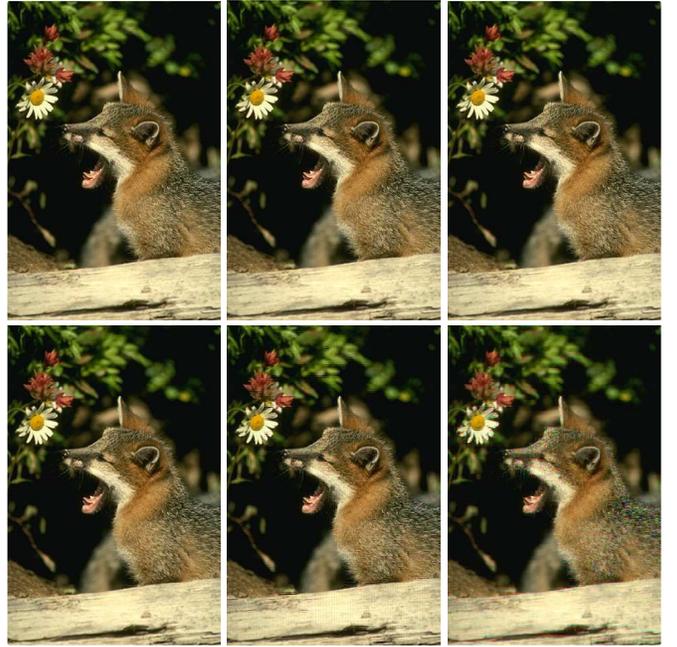}\caption{Reconstruction example when $M_{1}=M_{2}=6$. The images from left
to right, top to bottom and their PSNR (dB) values are: II+cTKSVD
(35.41), I+cTKSVD (34.97), Sapiro's+cKSVD (33.64), II+TKSVD (33.57),
SS+KHOSVD (28.62), Gaussian+cTKSVD (28.05). \label{fig:gallary}}
\end{figure}

\begin{figure}
\centering{}\includegraphics{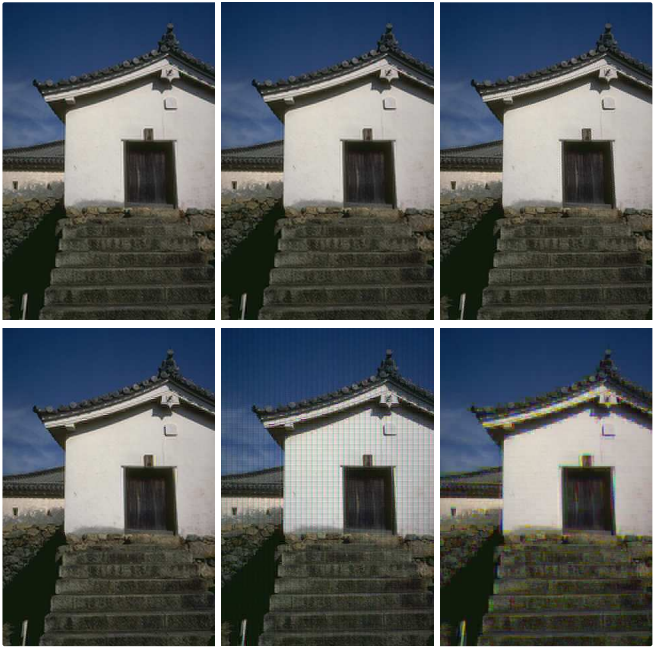}\caption{Reconstruction example when $M_{1}=M_{2}=4$. The images from left
to right, top to bottom and their PSNR (dB) values are: II+cTKSVD
(29.91), I+cTKSVD (29.45), Sapiro's+cKSVD (28.72), II+TKSVD (26.60),
SS+KHOSVD (22.62), Gaussian+cTKSVD (21.94). \label{fig:gallary-1}}
\end{figure}

The results are shown in Fig. \ref{fig:comb_M_sigma}. We can see
that the proposed approaches obtain higher PSNR values than all of
the other methods and II + cTKSVD performs best. To see the gain of
coupling sensing matrices during dictionary learning and optimizing
the sensing matrices, respectively, we compare II + cTKSVD with II
+ TKSVD and Gaussian + cTKSVD. For instance, when $M_{i}=5,\ \sigma^{2}=0$,
II + cTKSVD has a gain of about 3dB over II + TKSVD and nearly 9dB
over Gaussian + cTKSVD. Although Sapiro's + cKSVD has a similar performance
to ours at some specific settings, it is not for a TCS system that
requires multiple separable sensing matrices. Examples of reconstructed
images using these methods are demonstrated in Fig. \ref{fig:gallary}
and \ref{fig:gallary-1} with the corresponding PSNR values listed.
All of the conducted simulations verify that the proposed methods
of multidimensional sensing matrix and dictionary optimization improve
the performance of a TCS system.


\section{Conclusions}

\label{sec:conclusions}

In this paper, we propose to jointly optimize the multidimensional
sensing matrix and dictionary for TCS systems. To obtain the optimized
sensing matrices, a separable approach with closed form solutions
has been presented and a joint iterative approach with novel design
measures has also been proposed. The iterative approach certainly
has higher complexity, but also exhibits better performance. An approach
to learning the multidimensional dictionary has been designed, which
explicitly takes the multidimensional structure into account and removes
the redundant updates in the existing multilinear approaches in the
literature. Further gain is obtained by coupling the multidimensional
sensing matrix while learning the dictionary. The performance advantage
of the proposed approaches has been demonstrated by experiments using
both synthetic data and real images.

\appendices{}

\section{Proof of Theorem 3}

\label{sec:AppendixI}

Assume $\mathbf{A}_{i}=\boldsymbol{\Phi}_{i}\boldsymbol{\Psi}_{i}=\mathbf{U}_{\mathbf{A}_{i}}\left[\begin{array}{cc}
\boldsymbol{\Lambda}_{\mathbf{A}_{i}} & \mathbf{0}\end{array}\right]\mathbf{V}_{\mathbf{A}_{i}}^{T}$ is an SVD of $\mathbf{A}_{i}$ for $i=1,\ 2$ and $rank(\mathbf{A}_{i})=M_{i}$.
Then the objective we want to minimize in (\ref{eq:Approach I}) can
be rewritten as:

\textit{\footnotesize{}
\[
\left|\left|\mathbf{I}_{\hat{N}_{1}\hat{N}_{2}}-(\mathbf{V}_{\mathbf{A}_{2}}\left[\begin{array}{cc}
\boldsymbol{\Lambda}_{\mathbf{A}_{2}}^{2} & \mathbf{0}\\
\mathbf{0} & \mathbf{0}
\end{array}\right]\mathbf{V}_{\mathbf{A}_{2}}^{T})\otimes(\mathbf{V}_{\mathbf{A}_{1}}\left[\begin{array}{cc}
\boldsymbol{\Lambda}_{\mathbf{A}_{1}}^{2} & \mathbf{0}\\
\mathbf{0} & \mathbf{0}
\end{array}\right]\mathbf{V}_{\mathbf{A}_{1}}^{T})\right|\right|{}_{F}^{2}.
\]
}Denote $\boldsymbol{\Sigma}=\left[\begin{array}{cc}
\boldsymbol{\Lambda}_{\mathbf{A}_{2}}^{2} & \mathbf{0}\\
\mathbf{0} & \mathbf{0}
\end{array}\right]\otimes\left[\begin{array}{cc}
\boldsymbol{\Lambda}_{\mathbf{A}_{1}}^{2} & \mathbf{0}\\
\mathbf{0} & \mathbf{0}
\end{array}\right]=diag(\boldsymbol{\nu}_{\mathbf{A}_{2}}\otimes\boldsymbol{\nu}_{\mathbf{A}_{1}}),$ $\boldsymbol{\nu}_{\mathbf{A}_{i}}=diag(\left[\begin{array}{cc}
\boldsymbol{\Lambda}_{\mathbf{A}_{i}}^{2} & \mathbf{0}\\
\mathbf{0} & \mathbf{0}
\end{array}\right])$, then we have 
\begin{equation}
||\mathbf{I}_{\hat{N}_{1}\hat{N}_{2}}-(\mathbf{V}_{\mathbf{A}_{2}}\otimes\mathbf{V}_{\mathbf{A}_{1}})\boldsymbol{\Sigma}(\mathbf{V}_{\mathbf{A}_{2}}^{T}\otimes\mathbf{V}_{\mathbf{A}_{1}}^{T})||_{F}^{2}.\label{eq:F_norm}
\end{equation}
Let $\boldsymbol{\nu}_{\mathbf{A}_{i}}=[(v_{i})_{1},...,(v_{i})_{M_{i}},\mathbf{0}]^{T}$,
then the sub-vector of the diagonal of $\boldsymbol{\Sigma}$ containing
its non-zero values is: \textit{\small{}$\hat{\boldsymbol{\nu}}=[(v_{2})_{1}(v_{1})_{1},...,(v_{2})_{1}(v_{1})_{M_{1}},...,(v_{2})_{M_{2}}(v_{1})_{1},...,(v_{2})_{M_{2}}(v_{1})_{M_{1}}]^{T}$.}
Thus (\ref{eq:F_norm}) becomes: 
\begin{equation}
||\mathbf{I}_{\hat{N}_{1}\hat{N}_{2}}-\boldsymbol{\Sigma}||_{F}^{2}=\hat{N}_{1}\hat{N}_{2}-M_{1}M_{2}+\sum_{p=1}^{M_{2}}\sum_{q=1}^{M_{1}}(1-(v_{2})_{p}(v)_{q})^{2}.
\end{equation}
Therefore we can obtain that the minimum value of (\ref{eq:Approach I})
is $\hat{N}_{1}\hat{N}_{2}-M_{1}M_{2}$, and that it is achieved when
the entries of $\hat{\boldsymbol{\nu}}$ are all unity. 

Clearly $\boldsymbol{\Lambda}_{\mathbf{A}_{i}}$=$\mathbf{I}_{M_{i}}$
for $i=1,\ 2$ is a solution, i.e., $\mathbf{A}_{i}=\mathbf{U}_{\mathbf{A}_{i}}\left[\begin{array}{cc}
\mathbf{I}_{M_{i}} & \mathbf{0}\end{array}\right]\mathbf{V}_{\mathbf{A}_{i}}^{T}$ with $\mathbf{U}_{\mathbf{A}_{i}}\in\mathbb{R}^{M_{i}\times M_{i}}$
and $\mathbf{V}_{\mathbf{A}_{i}}\in\mathbb{R}^{\hat{N}_{i}\times\hat{N}_{i}}$
being arbitrary orthonormal matrices. Then we would like to find $\boldsymbol{\Phi}_{i}\ (i=1,\ 2)$
such that $\boldsymbol{\Phi}_{i}\boldsymbol{\Psi}_{i}=\mathbf{U}_{\mathbf{A}_{i}}\left[\begin{array}{cc}
\mathbf{I}_{M_{i}} & \mathbf{0}\end{array}\right]\mathbf{V}_{\mathbf{A}_{i}}^{T}$. Following the derivation of Theorem 2 in \cite{Li2013}, the solution
in (\ref{eq:solutionApproI}) can be found. 

With this solution, for an arbitrary vector $\mathbf{z}\in\mathbb{R}^{\hat{N}_{i}}$,
we have $||\mathbf{A}_{i}^{T}\mathbf{z}||_{2}^{2}=tr(\mathbf{z}^{T}\mathbf{A}_{i}\mathbf{A}_{i}^{T}\mathbf{z})=tr(\mathbf{z}^{T}\mathbf{z})=||\mathbf{z}||_{2}^{2}$,
which indicates that the resulting equivalent sensing matrices $\mathbf{A}_{i}\ (i=1,\ 2)$
are Parseval tight frames. In addition, we observe that the solution
in (\ref{eq:solutionApproI}) can be obtained by separately solving
the sub-problems in (\ref{eq:sub_prob_I}), of which the solutions
have been derived in \cite{Li2013}. By substituting the solutions
of the sub-problems into (\ref{eq:Approach I}), we can conclude the
minimum remains as $\hat{N}_{1}\hat{N}_{2}-M_{1}M_{2}$. 

\bibliographystyle{IEEEbib}
\bibliography{ProjectDesign,fyr,tensor}

\end{document}